\documentclass[journal]{IEEEtran}
\usepackage{amsmath}
\usepackage{url}
\usepackage{booktabs}
\usepackage{graphicx}
\usepackage{subcaption}
\usepackage{multirow}

\usepackage[pagebackref=true,breaklinks=true,letterpaper=true,colorlinks,bookmarks=false]{hyperref}

\newcommand{\etal}{\textit{et al.}}

\usepackage[switch]{lineno}
\usepackage{gensymb}
\usepackage{diagbox}

\begin{document}
%
\title{AU-Aware Vision Transformers \\ for Biased Facial Expression Recognition }
%
%
%

\author{Shuyi Mao,
        Xinpeng Li,
        Qingyang Wu,
        and~Xiaojiang Peng,~\IEEEmembership{Member,~IEEE}
\thanks{Shuyi Mao and Xinpeng Li are equally-contributted authors.}
\thanks{Corresponding Author: Xiaojiang Peng(pengxiaojiang@sztu.edu.cn)}}

\maketitle

\begin{abstract}
Studies have proven that domain bias and label bias exist in different Facial Expression Recognition (FER) datasets, making it hard to improve the performance of a specific dataset by adding other datasets. For the FER bias issue, recent researches mainly focus on the cross-domain issue with advanced domain adaption algorithms. This paper addresses another problem: how to boost FER performance by leveraging cross-domain datasets. Unlike the coarse and biased expression label, the facial Action Unit (AU) is fine-grained and objective suggested by psychological studies. Motivated by this, we resort to the AU information of different FER datasets for performance boosting and make contributions as follows. 
First, we experimentally show that the naive joint training of multiple FER datasets is harmful to the FER performance of individual datasets. We further introduce expression-specific mean images and AU cosine distances to measure FER dataset bias. This novel measurement shows consistent conclusions with experimental degradation of joint training. Second, we propose a simple yet conceptually-new framework, AU-aware Vision Transformer (AU-ViT). It improves the performance of individual datasets by jointly training auxiliary datasets with AU or pseudo-AU labels. We also find that the AU-ViT is robust to real-world occlusions. Moreover, for the first time, we prove that a carefully-initialized ViT achieves comparable performance to advanced deep convolutional networks. Our AU-ViT achieves state-of-the-art performance on three popular datasets, namely 91.10\% on RAF-DB, 65.59\% on AffectNet, and 90.15\% on FERPlus. The code and models will be released soon.

\end{abstract}

\section{Introduction}
\label{intro}
\IEEEPARstart{F}acial Expressions Recognition (FER) aims to identify people's emotions in a facial image. It plays a great role in various applications including surveillance \cite{clavel2008fear}, games \cite{pioggia2005android}, educations \cite{yang2018emotion}, medical treatments \cite{pioggia2005android} and marketing \cite{ren2012linguistic}. In the past decade, more and more scholars in computer vision have 
devoted themselves to this task and contributed a lot. \cite{corneanu2016survey, noroozi2018survey, rouast2019deep, zhang2018facial, li2020deep} Recently, many studies show one of the main challenges is data bias, which mainly comes from data composition \cite{li2020deeper, chen2021cross, panda2018contemplating} and label ambiguity (e.g. subjective annotation \cite{zeng2018facial, chen2021understanding} and compound expression \cite{li2017reliable}).

\begin{figure}[ht]
\centering
\includegraphics[width=0.5\textwidth]{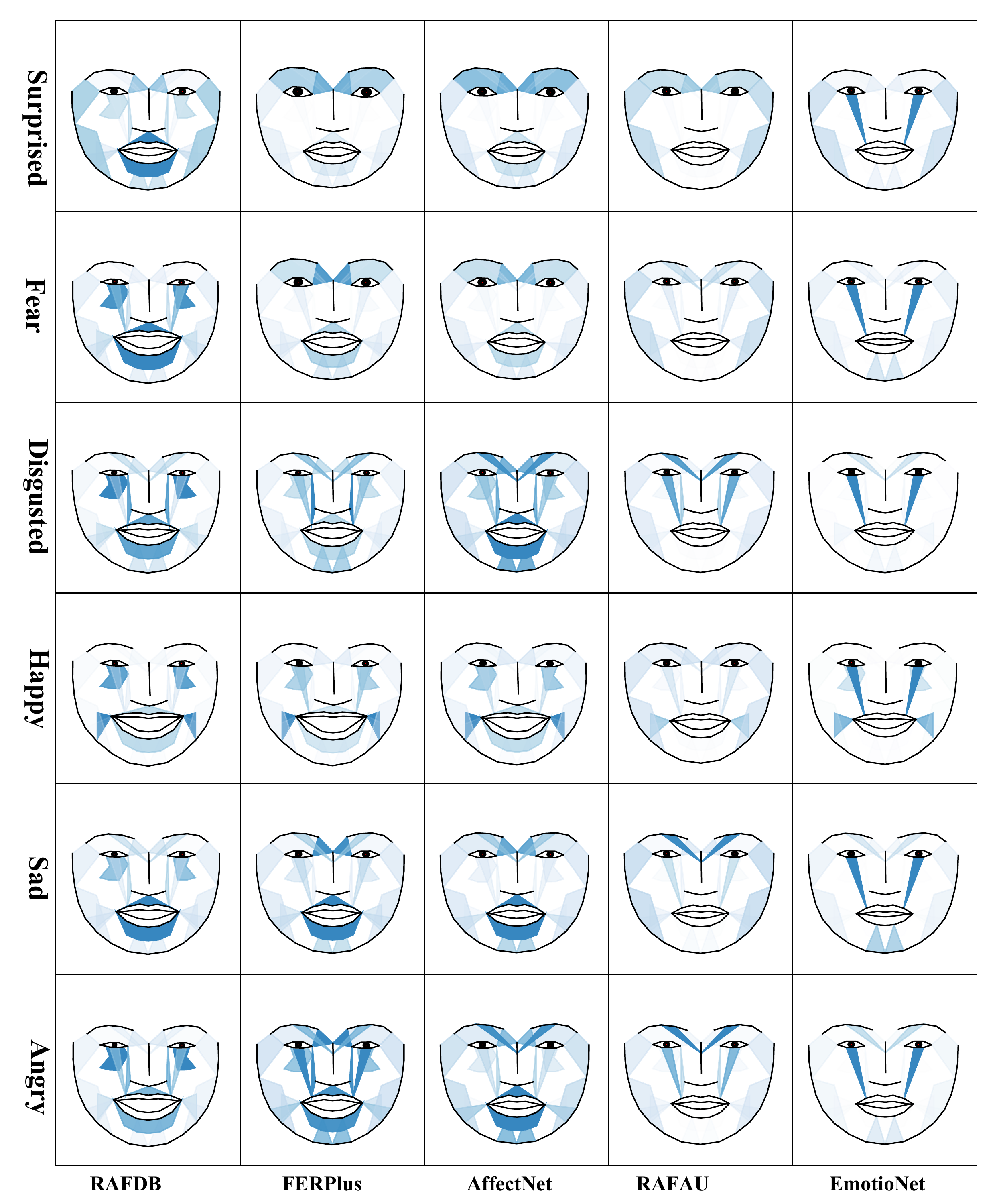}
\caption{The mean activated facial AUs of different emotions on popular FER datasets like RAFDB, FERPlus, AffectNet, RAFAU, and EmotioNet. The blue patches indicate AU regions and the color density refers to AU intensity.}
\label{fig:au}
\end{figure}

Dataset bias results in sub-optimal performance and even serious deterioration in joint dataset training. Existing works have achieved a lot in eliminating dataset bias, which can be mainly categorized into (1) label modification \cite{panda2018contemplating, zeng2018facial, wang2020suppressing, chen2020label, she2021dive, chen2021understanding} and (2) cross-domain adaptation \cite{li2020deeper, chen2021cross, zeng2022face2exp}. In label modification, the algorithms use web images, attention mechanisms, EM iteration, or distribution learning to correct biased labels. In cross-domain adaptation, the algorithms adopt regularization terms, adversarial learning, or meta mechanism to minimize domain divergence. However, they neither boost a target dataset's performance by leveraging auxiliary datasets nor bridge different datasets with objective information. 

Psychological studies show that Facial Action Unit (AU) is an objective and common standard to describe the physical expression of emotions \cite{ekman1997face}. For instance, $AU12$ and $AU6$ indicate lip corner puller and cheek raiser, which relate to happiness; $AU1$, $AU4$, and $AU6$ denote inner brow raiser, brow lowerer, and cheek raiser, which relate to sadness. Figure \ref{fig:au} shows the mean activated facial AUs (extracted by OpenFace~\cite{baltrusaitis2018openface} and drawn by Py-FEAT \footnote{https://github.com/cosanlab/py-feat}) of different emotions on several popular datasets. On the one hand, we observe that the AUs of the same expressions are inconsistent in different datasets, indicating dataset bias indirectly. On the other hand, since AUs are semantically universal among all datasets and partially shared in the same expression, we believe that the FER task can benefit from AUs.

In this paper, we first investigate the FER bias issue comprehensively. Existing works mainly suggest the bias existence by cross-dataset FER performance~\cite{li2020deeper, chen2021cross, zeng2018facial, panda2018contemplating}, i.e., training on a source dataset and testing on another target dataset. Here, we mainly focus on boosting individual dataset performance with cross-domain datasets. We analyze dataset bias in a fine-grained view by regarding FER as an objective image-to-AU plus a subjective AU-to-emotion procedure. The bias of the image-to-AU process is more likely to be collection bias, and the one of the AU-to-emotion procedure is mainly about annotation bias due to subjectivity. We focus on measuring the later procedure since the large-scale data and deep neural network are expected to eliminate the collection bias. Specifically, we use AU differences to estimate the annotation subjectivity qualitatively and quantitatively. In the quantitative aspect, we show that the performance degrades for a specific dataset by joint training with other FER datasets. We plot the expression-specific mean images and AU cosine distance 
among datasets in the qualitative aspect. The cosine distance is consistent with joint training performance degradation. 

Second, we propose a conceptually-new yet simple network architecture, termed an AU-aware Vision Transformer (AU-ViT), to jointly train a target dataset and cross-domain datasets, which can effectively boost the performance of the target dataset. Since the image-to-AU procedure is objective, the AU-ViT mainly takes AUs or pseudo-AUs as a bridge between target and auxiliary datasets. The AU-ViT consists of three vital branches: a Base branch, a ViT-based Expression classification branch (ViT-Exp branch), and an AU branch. For the Base branch, two alternatives are designed, i.e., vanilla ViT, which conducts Transformers on pixel patches, and convolution-based ViT (CNN-ViT), which applies Transformers on feature maps. Besides, we empirically introduce advanced modules like multi-stage blocks. For the AU branch, we elaborately design several patch-splitting strategies for feature aggregation, motivated by the fact that AUs are mainly defined by local information. Plus, we introduce a symmetric maxout layer to address occlusion and profile issues. The AU branch is expected to boost the Base branch's feature learning and thus lead to better expression recognition. 

Finally, we extensively evaluate our approach on three popular FER datasets, two AU datasets, and three occlusion FER test datasets. Our AU-ViT achieves state-of-the-art FER performance with 91.10\% on RAF-DB, 65.59\% on AffectNet, and 90.15\% on FERPlus. The AU-ViT also outperforms current state-of-the-art methods on Occlusion-RAFDB, Occlusion-FERPlus, and Occlusion-FERPlus with clear margins. 
In addition, we find that a vanilla ViT pipeline with proper initialization (Base branch plus ViT-Exp branch) achieves comparable performance to advanced deep convolutional networks. With visualizations, we observe that our AU-ViT captures more meaningful semantic features than other methods. The code and pre-trained models will be released soon.

Our contributions can be summarized as follows:
\begin{itemize}
  \item [1)] We conduct extensive FER datasets joint training experiments. We find naive joint training seriously decreases the FER performance of a particular dataset. Then we measure FER dataset bias in a fine-grained way by computing expression-specific mean images and AU cosine distances. The measurement shows consistent conclusions with experimental degradation. 
  \item [2)] We propose a novel AU-aware Vision Transformer (AU-ViT) architecture to train a target dataset and cross-domain datasets jointly. The AU branch of AU-ViT resorts to AU information and effectively boosts the performance of the target dataset. We also introduce several patch-splitting schemes for the AU branch. Visualizations show that our AU-ViT captures more meaningful semantic features than other methods due to the usage of AUs.
  \item [3)] To the best of our knowledge, This is the first work to repurpose a vanilla ViT pipeline for FER. We demonstrate that a carefully initialized vanilla ViT obtains comparable performance to advanced deep convolutional networks. Besides, the CNN embedding and advanced modules like multi-stage blocks further boost the performance.
  \item [4)] We conduct extensive experiments on eight publicly available FER and AU databases to evaluate our AU-ViT and achieve state-of-the-art performance with 91.10\% on RAF-DB, 65.59\% on AffectNet, and 90.15\% on FERPlus.
\end{itemize}


\section{Related work}
\label{related}
Numerous scholars in computer vision have devoted themselves to FER and contributed a lot of works, including algorithms~\cite{1962Method, 2006Haar, 2009Facial, corneanu2016survey, noroozi2018survey, rouast2019deep, zhang2018facial, li2020deep, Farzaneh_2020_CVPR_Workshops, zhao2021robust, wang2020suppressing, wang2020region} and datasets~\cite{ZHANG2014692, 2010The, li2017reliable, BarsoumICMI2016}. This paper 
proposes AU-ViT to address how to boost FER performance by leveraging cross-domain datasets. Therefore, we will show related works closely related to this issue and methods, including cross-domain FER, joint training FER, dataset bias in FER, vision transformers in FER, and AU assistance in FER. 

\textbf{Dataset bias in FER}.
Dataset bias exists among different FER datasets due to data composition~\cite{li2020deeper, chen2021cross, panda2018contemplating} and label ambiguity (e.g. subjective annotation~\cite{zeng2018facial, chen2021understanding} and compound expression~\cite{li2017reliable}). Existing works have achieved a lot in eliminating dataset bias. Wang \etal~\cite{wang2020suppressing} modify a pair of training labels with the self-attention mechanism. Chen \etal~\cite{chen2020label} leverage action unit recognition and facial landmarks detection to guide label distributions. She \etal~\cite{she2021dive} build a multi-branch framework to mine label distribution and uncertainty extent in label space. \textit{Different from these works, we resort to the AU information of different FER datasets for bias measurement}.

\textbf{Joint training and cross-domain in FER}.
Some works have explored joint training in FER to improve algorithm generalization. Pan \etal~\cite{panda2018contemplating} collect web images for more diversified image-label pairs. Zeng \etal~\cite{zeng2018facial} assign each sample several labels and find the latent truth in joint training. 

Recent methods regard skipping dataset bias as a cross-domain problem. Li \etal~\cite{li2020deeper} propose to minimize an emotion-conditional maximum mean discrepancy to learn domain-invariant features. Chen \etal~\cite{chen2021cross} adopt holistic and local features between datasets with an adversarial graph representation learning. Zeng \etal~\cite{zeng2022face2exp} utilize face recognition datasets to address FER's class imbalance with an additional adaptation network. \textit{Different from these works, we address how to boost FER performance of a target dataset by leveraging cross-domain datasets}.

\textbf{Vision Transformer in FER}.
Vision Transformers have been proved effective in image classification, detection, and segmentation~\cite{dosovitskiy2020image, 2021CrossViT, wang2021pyramid, heo2021rethinking, liu2021swin}. Recently, some works introduced Transformers to the FER task. Ma \etal~\cite{ma2021facial} combine two kinds of feature maps generated by two-branch CNNs, then feed the fused feature into Transformers to explore relationships between visual tokens. Xue \etal~\cite{xue2021transfer} leverage Transformers upon CNNs to learn rich relations among diverse local patches between two attention droppings. \textit{Different from these works, we build an AU-aware branch for patches after Transformer, which guides the blocks of Transformer to learning finer facial expression features}.

\textbf{AU assistance in FER}.
Previous methods utilize AU information for FER since AUs are related to emotions~\cite{ekman1978facial}. Deng \etal~\cite{deng2020multitask} present a distillation strategy to learn from incomplete labels to boost the performance of AU detection, expression classification, and valence-arousal estimation. Pu \etal~\cite{pu2021expression} exploit the relationships between AU and expression and choose useful AU representations for FER. Chen \etal~\cite{chen2021understanding} leverage AUs to understand and mitigate annotation bias and adopt triplet loss to encourage embeddings with similar AUs to get close in feature space. \textit{Different from these works, we leverage AU or pseudo-AU labels from any other facial datasets and embed them into Transformers, aiming to boost the accuracy of target datasets}.


\section{Methodology}
\label{method}
To bridge the dataset bias and boost the accuracy of the target dataset, we propose an AU-aware Vision Transformer (AU-ViT) for multiple datasets training in FER. We first show the FER dataset bias qualitatively and quantitatively and then detail the components of the AU-ViT.

\subsection{Dataset bias}
Extensive datasets usually contribute to better network training in the deep learning era. However, the performance of a specific dataset degrades when training with auxiliary FER datasets due to bias. The existing works measure FER bias by dataset recognition and cross-dataset generalization~\cite{zeng2018facial, panda2018contemplating, chen2021cross, li2020deeper}. Here, we analyze FER bias in a fine-grain view, which regards FER as an image-to-AU and AU-to-expression procedure. The image-to-AU process mainly contains expression-independent bias (e.g., illumination), while the AU-to-expression procedure includes expression-dependent discrimination. We focus on the latter bias since the former is expected to be eliminated by large-scale datasets or robust deep convolutional networks, which is less related to joint training degradation.

\begin{figure}[!t]
\centering
\includegraphics[width=0.5\textwidth]{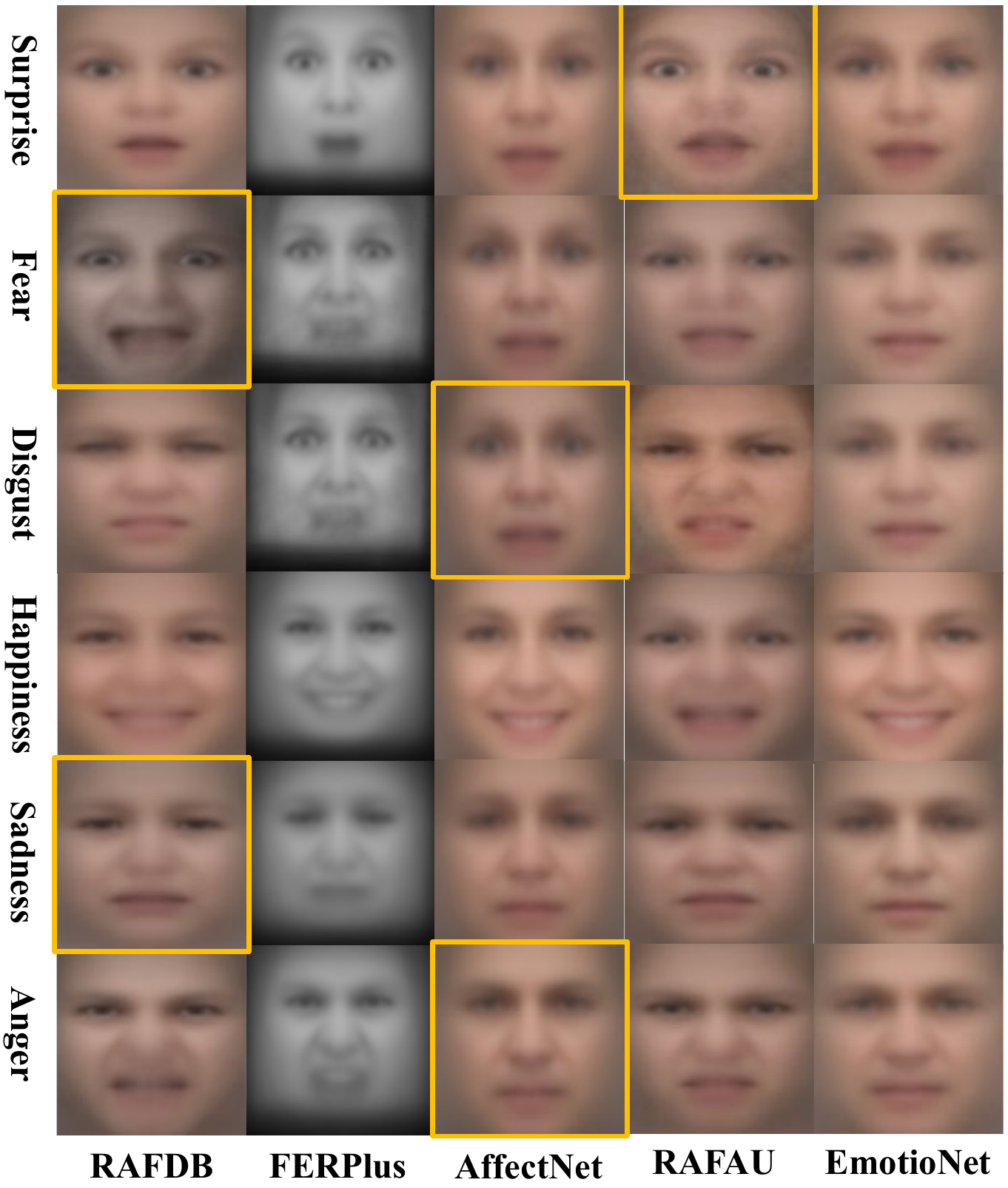}
\caption{Illustration of the AU-to-emotion Bias. We calculate the mean images of different emotions and datasets. The yellow box indicates the image with different facial motions.}
\label{fig: AU-to-emotion bias}
\end{figure}

\textbf{Dataset mean images.}
We show the mean images of different expressions among different datasets in Figure \ref{fig: AU-to-emotion bias}. For comparison, we crop all images into the face image ratio of RAFDB and resize them into 112 x 112 size. Then, we calculate the mean images, which encode the statistical AUs of expressions in a specific dataset. We can see different relations between AUs and expressions among different datasets from the mean pictures, which indicate the dataset bias qualitatively.

As shown in Figure \ref{fig: AU-to-emotion bias}, the mean images of the same expressions vary more or less in different datasets. For example, the surprise average image of FERPlus shows more vigorous facial motion than one of the other datasets. The difference results from intra-class variance and annotation subjectivity. Worth mentioning that happiness shows more consistent facial movement in various datasets than other expressions.

\begin{figure*}[!t]
\centering
\includegraphics[width=1\textwidth]{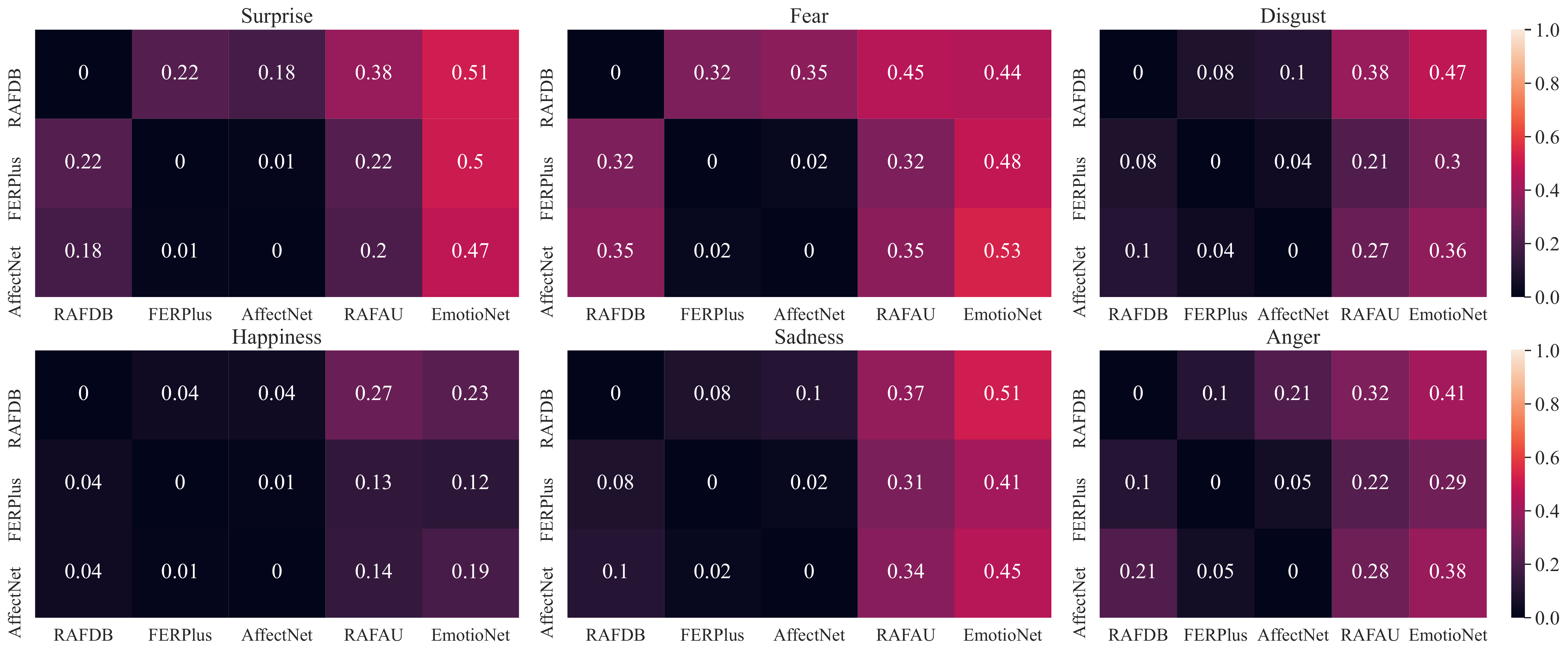}
\caption{Illustration of the quantitative AU-to-expression bias. We calculate the mean AU vectors of all emotions in different datasets and then plot their cosine distance. It is worth mentioning that two vectors are regarded as similar enough when the distance is less than $0.3$ and different enough when more than $0.5$. }
\label{fig: cosine distance}
\end{figure*}

\renewcommand\arraystretch{2}
\begin{table*}[ht]
\center

\begin{tabular}{c||c|c|c|c|c|r}
\hline
\diagbox{Target}{Accuracy (\%)}{Auxiliary}         & RAFDB & FERPlus & AffectNet & RAFAU & EmotioNet & None \\ \hline\hline
RAFDB    & /      & 84.84 (-2.46)   & 86.67 (-0.63)    & 82.72 (-4.58)      & 85.10 (-2.2)         & \textbf{87.3}     \\ \hline
FERPlus   & 86.54 (-1.37)  & /       & 87.69 (-0.22)    & 87.19 (-0.87)     & 86.93 (-0.98)         & \textbf{87.91}    \\ \hline
AffectNet & 58.19 (-1.86) & 59.91 (-0.41)   & /         & 59.24 (-0.81)     & 58.88 (-1.17)         & \textbf{60.05}    \\
\hline
\end{tabular}

\caption{The results of joint training on dataset pairs with FER labels. The results are reported on target datasets. The bracket presents the performance degradation compared to traditional training on individual datasets (i.e., the last column).}
\label{tab:FER-FER}
\end{table*}




\textbf{Dataset cosine distance.}
To get a further accurate measurement for dataset bias, we plot the cosine distance of expression-specific facial AU representations between datasets in Figure \ref{fig: cosine distance}. Specifically, we adopt AU labels from dataset annotations or OpenFace predictions. For comparison, we choose commonly available AUs: AU01, AU02, AU04, AU05, AU06, AU09, AU10, AU12, AU15, AU17, AU20, AU25, and AU26. Therefore, a 13-dimension mean AU vector represents the facial information of an expression in a dataset. Then, we compute their cosine distance, ranging from $0.0$ to $1.0$, to show dataset bias. It is worth mentioning that two vectors are regarded as similar enough when the space is less than $0.3$ and different enough when more than $0.5$.

From a view of general expressions, the EmotioNet is the most different dataset and contains around $0.5$ distance from others. The distance between the RAFDB, FERPlus, and AffectNet is less than $0.3$. From a view of a specific expression, fear is the most contrasting expression, and it includes more than $0.3$ distance from other datasets. Happiness is the most consistent expression and holds less than $0.2$ distance. These biases are constant with the following performance degradation of joint-training results.

\textbf{Joint training exploration.}
We train different dataset pairs jointly to show how dataset bias affects the performance of FER. Specifically, we choose from three FER datasets: RAFDB, FERPlus, and AffectNet, and two AU datasets: RAFAU and EmotioNet. We use the  deepface\footnote{https://github.com/serengil/deepface}~\cite{serengil2020lightface, serengil2021lightface} to predict expressions and OpenFace~\cite{baltrusaitis2018openface} to predict AUs if labels are unavailable in the datasets. To keep the dominated role of target FER datasets, we set the mixing ratio of target and auxiliary images as $4:1$. We pretrain the pure ViT on VGGFace2 in experiments and present the results in Table \ref{tab:FER-FER}.

From Table \ref{tab:FER-FER}, we can see that the results of joint training methods are consistently inferior to traditional training methods. For example, the accuracy of RAFDB drops by 2.46\%, 0.63\%, 4.58\%, and 2.2\% when combining FERPlus, AffectNet, RAFAU, and EmotioNet, respectively. Joint training with RAFAU results in the largest performance degradation for RAFDB, corresponding to the most significant quantitative difference in Figure \ref{fig: cosine distance}. Besides, the performance of FERPlus decreases by 1.37\%, 0.22\%, 0.87\%, and 0.98\% when combining RAFDB, AffectNet, RAFAU, and EmotioNet, respectively. 
We also find that joint training FERPlus with AffectNet only suffers a slight degradation, consistent with the small quantitative difference in Figure \ref{fig: cosine distance}. These results demonstrate that joint training with expression labels is not helpful in FER due to dataset bias and suggest the effectiveness of our quantitative expression bias scheme.


\begin{figure*}[!t]
\centering
\includegraphics[width=1\textwidth]{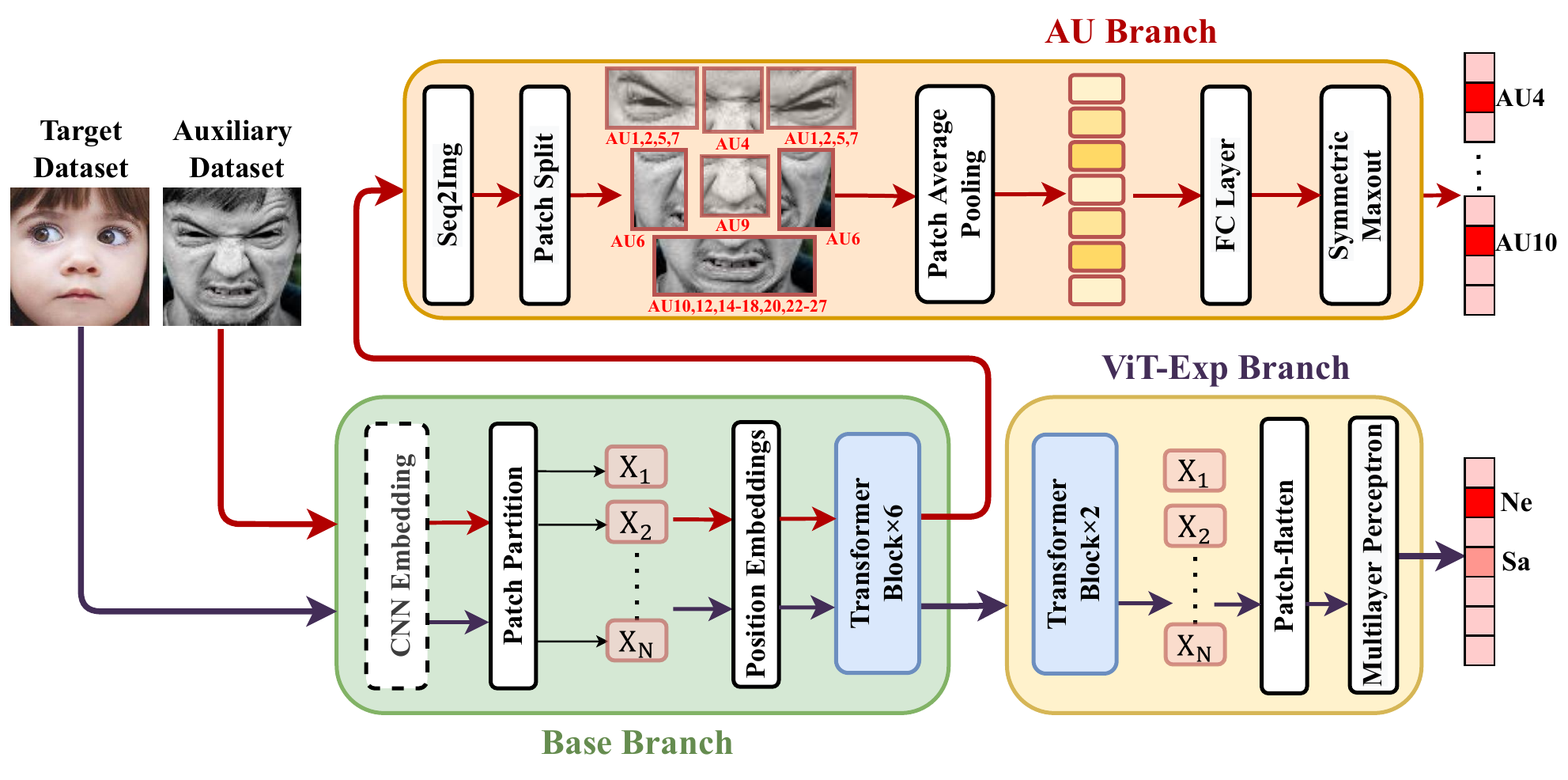}
\caption{Overview of our AU-ViT. It consists of three essential branches: Base branch, ViT-Exp branch, and AU branch. Specifically, the AU-ViT first takes as input the images from both target and auxiliary datasets, then encodes them into middle-level feature maps by the Base branch, and finally feeds the feature maps into either the ViT-Exp branch for facial expression classification or the AU branch for AU recognition. The CNN Embedding part is optional in the Base branch.}
\label{fig: pipeline}
\end{figure*}

\subsection{ AU-aware Vision Transformer}
From the above analysis, dataset bias is the critical problem for performance boosting in joint training. Existing methods mainly eliminate the dataset bias from an image-to-expression view, e.g., changing labels or regularizing embeddings. In this paper, regarding FER as an objective image-to-AU procedure plus a biased AU-to-expression procedure, we resort to the AU objectivity yet avoid the expression label bias to boost the performance of individual target datasets. Specifically, we propose the AU-ViT model for jointly training a target FER dataset and auxiliary datasets with their AUs or pseudo-AUs.

\textbf{AU-ViT Overview.} 
As shown in Figure \ref{fig: pipeline}, the AU-ViT consists of three essential branches, namely the Base branch, ViT-Exp branch, and AU branch. The AU-ViT first takes the images from both target and auxiliary datasets as input. The Base branch then encodes images into middle-level feature maps and feeds them into either the ViT-Exp branch for facial expression classification or the AU branch for AU recognition. Specifically, we sample the images from target and auxiliary datasets with a 4:1 ratio. The ViT-Exp branch only handles the target dataset, and the AU branch only deals with the auxiliary datasets which own AU or pseudo-AU labels. 

\textbf{Base branch.}
The Base branch is designed into two alternatives, i.e., the vanilla ViT and CNN-ViT. Note that both ViTs of the Base branch exclude the classification part of a standard ViT pipeline. For the vanilla ViT, we split the image into $N$ patches with $16\times 16$ size and then reshape them into a 256-dimension embedding $X$. Then, we combine $X$ and a learnable position embedding $P$ through element-wise addition. For the CNN-ViT, we adopt feature maps from the third stage of IR50~\cite{wang2021face} and conduct the above operations. The feature maps are split into non-overlapping patches with $1\times 1$ size, and the dimension of each patch embedding is 256. Note that we concatenate a class token with $X$ in the vanilla ViT while discarding it in the CNN-ViT. The above operations of can be formulated as follows,

\begin{equation}
Z = [X_{1};X_{2};...;X_{N}] + P,
\end{equation}
where $Z$ is the sequence to be fed into the Transformers. 

\begin{figure}[!t]
\centering
\includegraphics[width=0.5\textwidth]{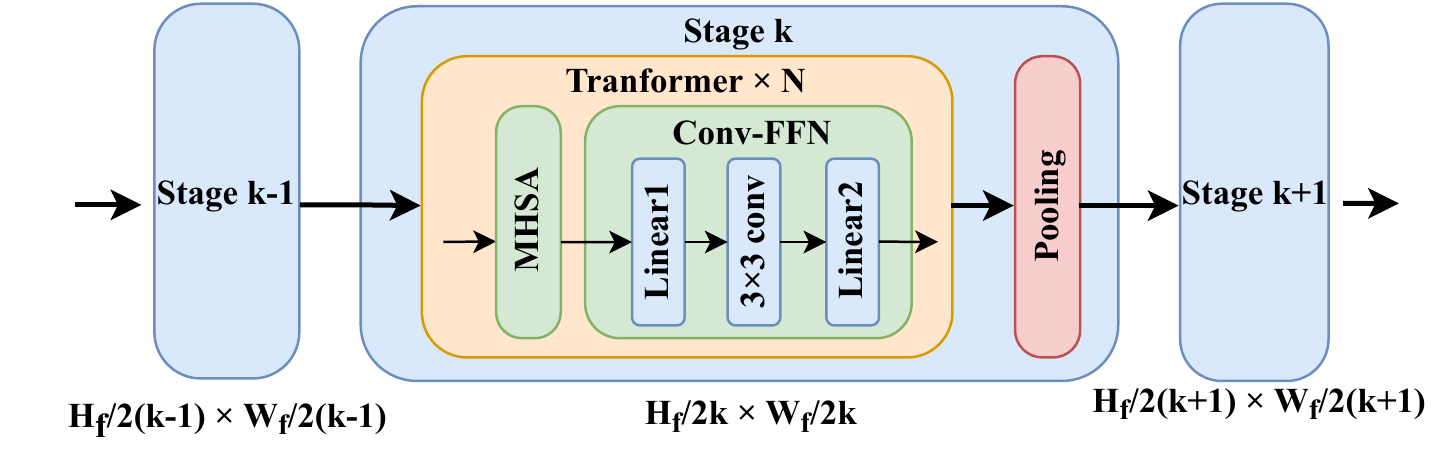}
\caption{Illustration of Multi-stage Transformers. The feature maps from the former stage are fed into MHSA, Conv-FFN, and a pooling layer. The size of feature map in the end of the k-th stage of Transformer is $\frac{H_{f}}{2k}\times \frac{W_{f}}{2k}\times 2kC_{f}$, where $H_{f}$, $W_{f}$, and $C_{f}$ are the height, width and dimension. }
\label{fig:Emotion Branch}
\end{figure}

Existing works show the Multi-stage Transformer performs better than primitive ones in computer vision task~\cite{wang2021pyramid, heo2021rethinking, liu2021swin}. To this end, we adopt it to build AU-ViT blocks experimentally. As shown in Figure \ref{fig:Emotion Branch}, Multi-stage Transformers gradually increase the channel size while decreasing the spatial size. Moreover, the number of tokens is reduced by patch merging layers as the network gets deeper. At the end of each stage, the pooling layer concatenates the tokens of each $2\times2$ neighbor patch and doubles their dimensions. Therefore, the size of feature map in the end of the k-th stage of Transformer is $\frac{H_{f}}{2k}\times \frac{W_{f}}{2k}\times 2kC_{f}$, where $H_{f}$, $W_{f}$, and $C_{f}$ are the height, width and dimension of the input feature map. 

Inside the Multi-stage Transformer, a Multi-Head Self-Attention (MHSA) module processes the input embedding and models the complex interactions. Then, a Convolution in Feed-Forward Network (Conv-FFN) is employed to capture the connectivity of local regions. In the Conv-FFN module, we rearrange the sequence of tokens into a 2D lattice and convert it to 2D feature maps. The number of channels of the feature maps increases firstly, and then a depth-wise convolution with a kernel size of $3\times3$ performs on them. Finally, we restore the channels of feature maps and flatten them into sequences with the initial dimension. The local convolution enhances the representation correlation with neighboring eight tokens. 

\textbf{ViT-Exp branch.}
The ViT-exp branch handles the middle-level feature maps of a target dataset from the Base branch. 
For the vanilla ViT base branch, we implement a Multi-Layer Perception on the $CLS$ Token to obtain the final classification. 
For the CNN-ViT base branch, two more Multi-stage Transformer blocks are further used upon the feature maps. Then, we flatten all the patch embeddings in the Patch-flatten layer and implement a Multi-Layer Perception for expression classification. Compared to pooling, the flattened operation and fully connected layer can retain all spacial information and the strong relationships among the features.

\textbf{AU branch.}
The AU branch processes the middle-level feature maps of the auxiliary dataset from the Base branch. A Seq2Img operation~\cite{liu2021swin} is first used to transform the patch tokens into 2D feature maps. Then, we crop the 2D feature maps into several overlapped patches according to AUs.

The top part of Figure \ref{fig: pipeline} plots our AU-patch splitting strategy. We split the feature maps into seven patches: the left eye, right eye, left cheek, right cheek, between-eyebrow, nose, and mouth. The left and right-eye patches are responsible for AU1, AU2, AU5, and AU7. Likewise, the left and right-cheek patches are both responsible for AU6. The between-eye and nose patches are in charge of AU4 and AU9, respectively. In particular, the mouth patch is the largest part among these patches, 
which is related to 14 kinds of AUs: AU10, AU12, AU14, AU15, AU16, AU17, AU18, AU20, AU22, AU23, AU24, AU25, AU26, AU27. After splitting the feature maps, we conduct average pooling in each patch (Patch Average Pooling in Figure \ref{fig: pipeline}) to get patch representations and feed them into the fully-connected layers for AU recognition. 

It is worth noting that the left and right patches, e.g., eye and cheek, are responsible for the same AUs. However, the occlusion and profile may lead to an asymmetric face and thus inconsistent AU prediction. To address this issue, we introduce a \textit{symmetric maxout} layer to select the maximum of the left and right prediction. The formulation is as follows,
\begin{equation}
z = max(z_{left}, z_{right}),
\end{equation}
where $z_{left}$ and $z_{right}$ denote the AU predictions of the left and right face, and $z$ is the final AU prediction.

\textbf{Loss Function}.
We adopt the cross-entropy loss ($L_{ce}$) for FER and binary cross-entropy loss ($L_{bce}$) for AU recognition. Let $z_{fer}$ and $z_{au}$ be the final output of the ViT-Exp branch and the AU branch, $y_{fer}$ and $y_{au}$ be the ground truth labels of the FER dataset and AU dataset, respectively. The former is fed into the softMax function while the latter inputs the Sigmoid function. The joint loss of AU-ViT is as follows:

\begin{equation}
\begin{split}
L &= \alpha L_{\text{ce}}(softmax(z_{fer}), y_{fer}) \\
&+ \beta L_{\text{bce}}(sigmoid(z_{au}), y_{au}),
\end{split}
\label{eq:loss}
\end{equation}
where $\alpha$ and $\beta$ are hyper-parameters to balance two tasks.


\section{Experiments}
\label{experiment}
In this section, we first describe the implementation details and datasets. Then we show quantitative and qualitative experimental results of eight datasets, which demonstrate the superiority of the AU-ViT in joint training. Finally, we conduct ablation studies of advanced modules and compare the AU-ViT with state-of-art methods.

\subsection{Implementation details}

The vanilla ViT is pre-trained on VGGFace2~\cite{cao2018vggface2} with an image size of 224$\times$224. The CNN Embedding module is initialized by IR50~\cite{wang2021face} pre-trained on MS1M~\cite{guo2016ms}. We generate pseudo-AU labels for the FER datasets by the AU detector OpenFace~\cite{baltrusaitis2018openface}. All the images are aligned and resized to 112$\times$112. For data augmentation, we use Random-Horizontal Flip, Random-Grayscale, and Gaussian Blur for all datasets. We use Adam optimizer for RAFDB and FERPlus, and SGD optimizer for AffectNet. The weight decay is 5e-4, and the batch size is 64 for RAFDB and 128 for FERPlus and AffectNet. The learning rate is initialized as 0.0001, and we use a linear learning rate warm-up of 10 epochs and a cosine learning rate decay of 10 epochs. All the experiments are implemented by Pytorch with 4 NVIDIA V100 GPUs. The default $\alpha$ and $\beta$ in Equation (\ref{eq:loss}) are 1.0 and 1.0, respectively.

\renewcommand\arraystretch{2}
\begin{table*}[ht]
\center
\begin{tabular}{c||c|c|c|c|c}
\hline
\diagbox{Target}{Accuracy (\%)}{Auxiliary}         & RAFDB & FERPlus & AffectNet & RAFAU & EmotioNet \\ \hline\hline
RAFDB  & /  & 88.41 (+3.57)    & 87.84 (+1.17)  & \textbf{88.8} (+6.08)   & 87.9 (+2.8)     \\ \hline
FERPlus  & 88.4 (+1.86)  & /  & 88.34 (+0.56)     & 88.44 (+1.25)  &\textbf{88.53} (+1.6)      \\ \hline
AffectNet & 61.57 (+3.38)  & 62.03 (+2.12)  & /  & \textbf{62.33} (+3.09)  & 61.17 (+2.83)  \\
\hline
\end{tabular}
\caption{The results of joint training on dataset pairs with our basic AU-ViT. The numbers in brackets present the improvements of our basic AU-ViT compared to the results of ViT in Table \ref{tab:FER-FER}.}
\label{tab:FER-AU}
\end{table*}

\begin{figure*}[t]
\centering
\includegraphics[width=0.9\textwidth]{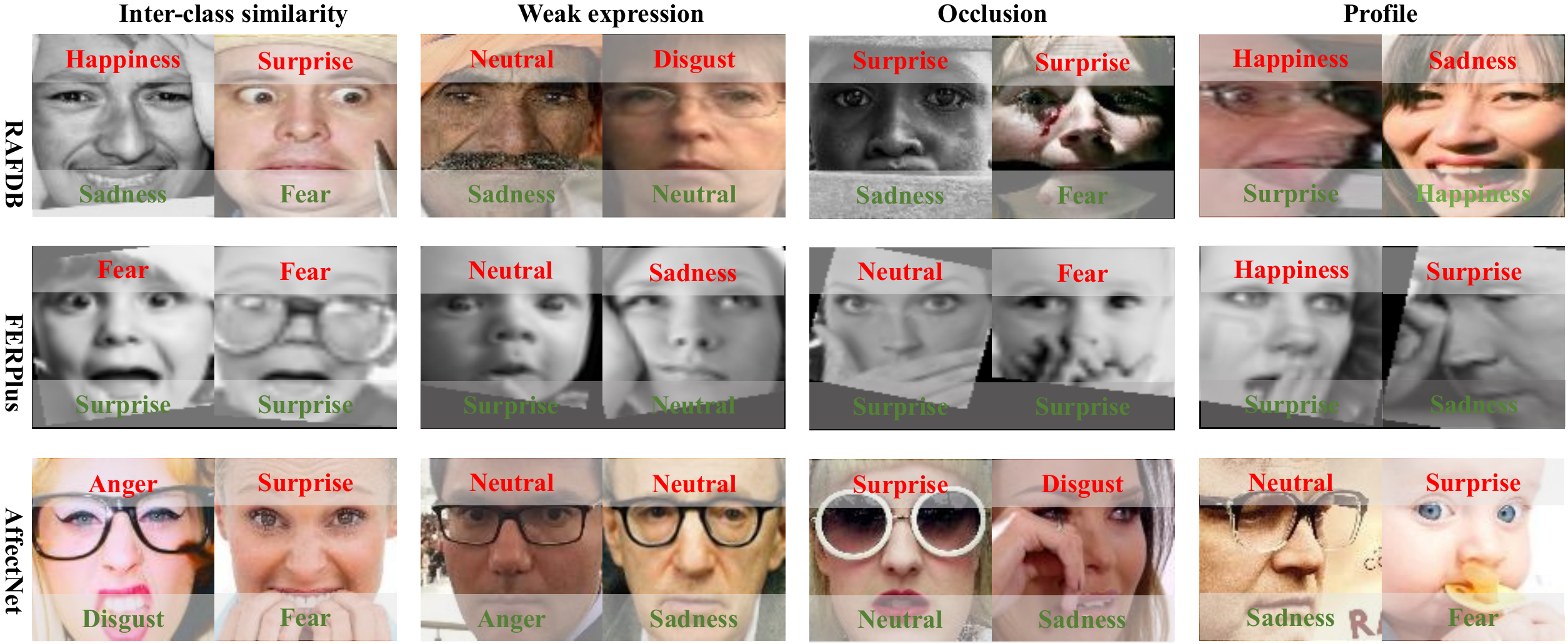}
\caption{The illustration of predicted samples by vanilla ViT trained on individual datasets and our basic AU-ViT jointly-trained with RAFAU. The red texts indicate incorrect predictions by vanilla ViT while the green ones indicate correct predictions by the AU-ViT. We can see the AU-ViT performs well in inter-class similarity, weak expression, occlusion, and profile situations.}
\label{fig: Wrong classed}
\end{figure*}

\begin{figure*}[t]
\captionsetup[subfigure]{labelformat=empty}
\begin{subfigure}{.33\textwidth}
  \centering
  \includegraphics[width=1\linewidth]{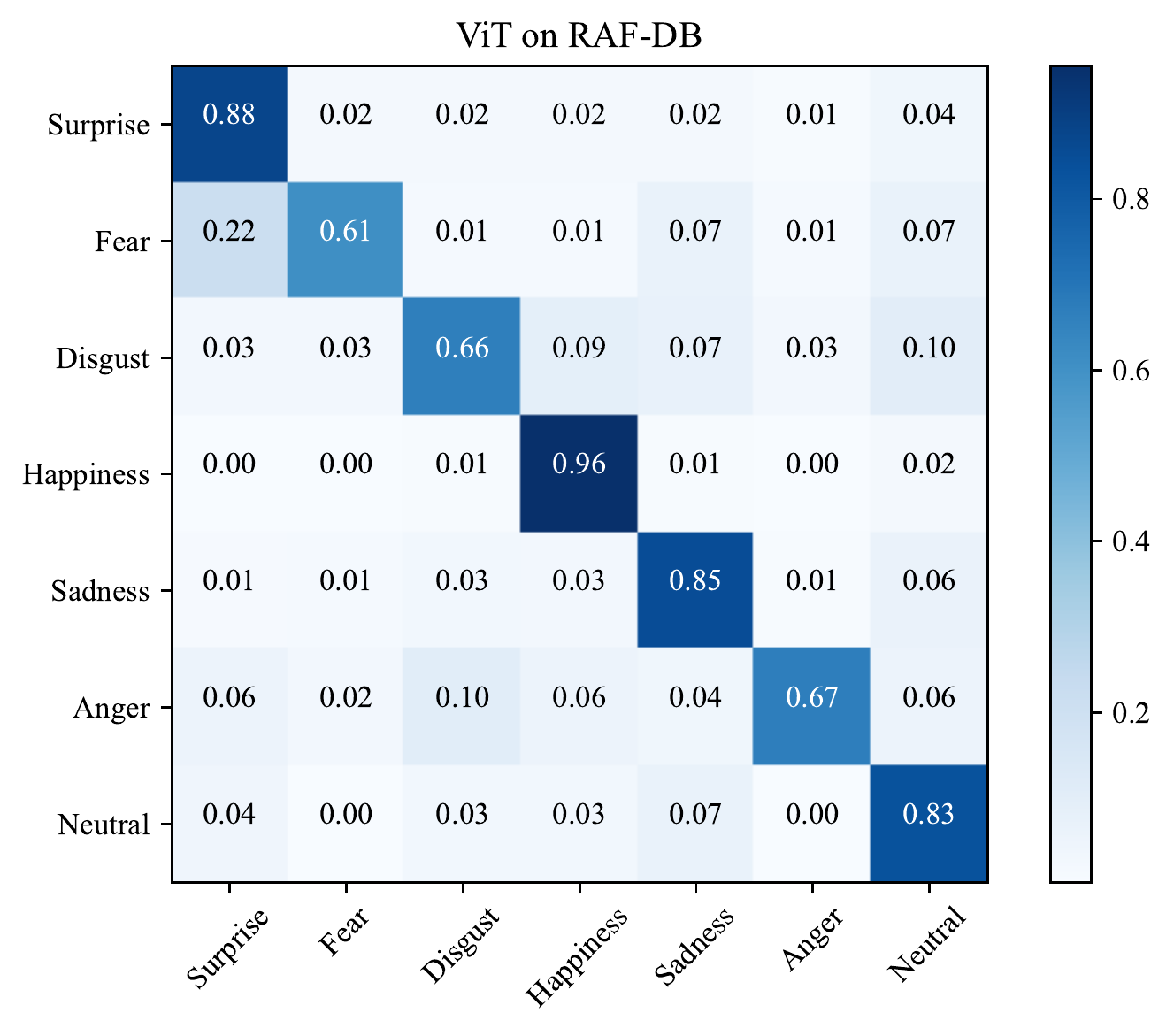}  
\end{subfigure}
\begin{subfigure}{.33\textwidth}
  \centering
  \includegraphics[width=1\linewidth]{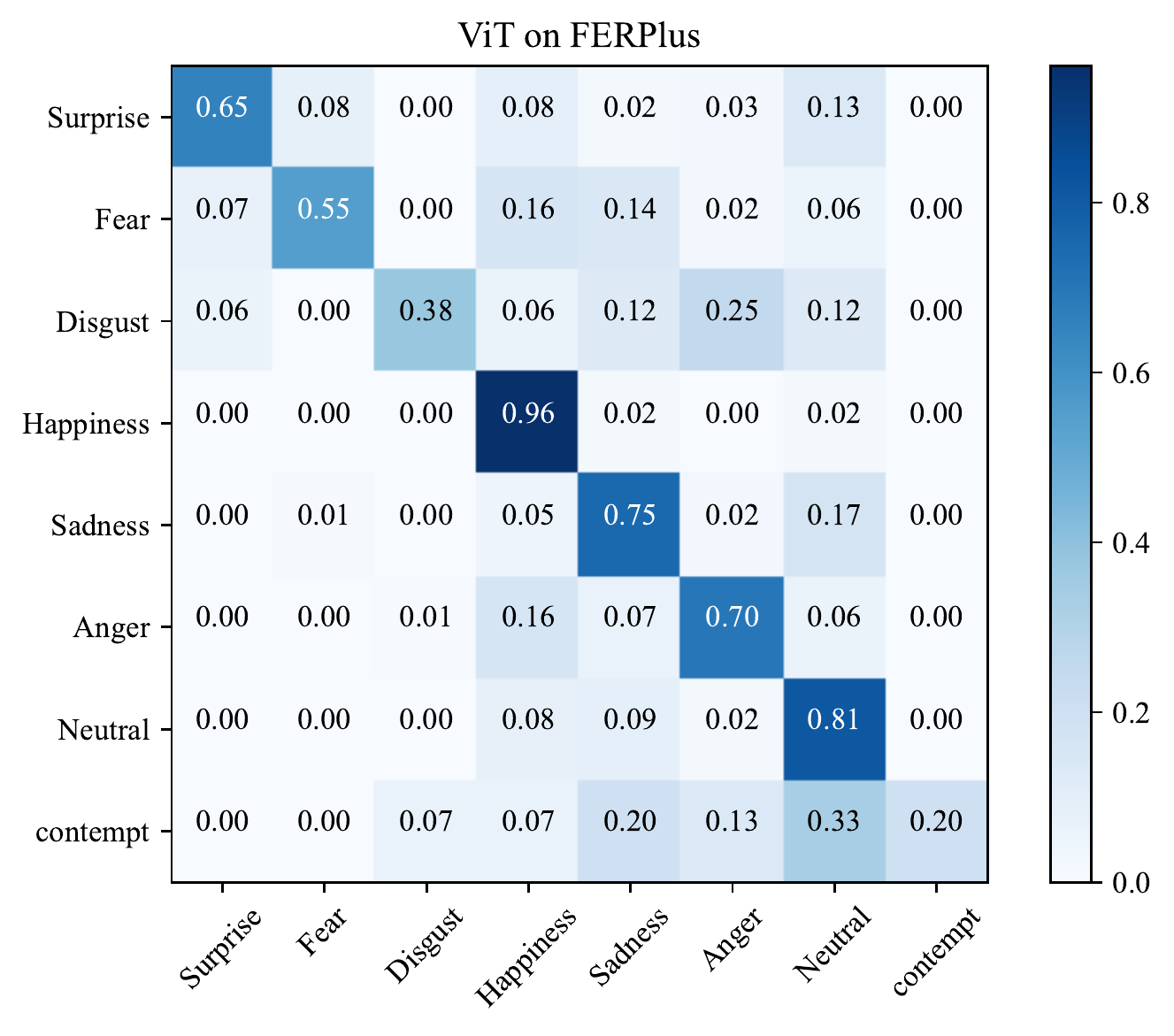}  
\end{subfigure}
\begin{subfigure}{.33\textwidth}
  \centering
  \includegraphics[width=1\linewidth]{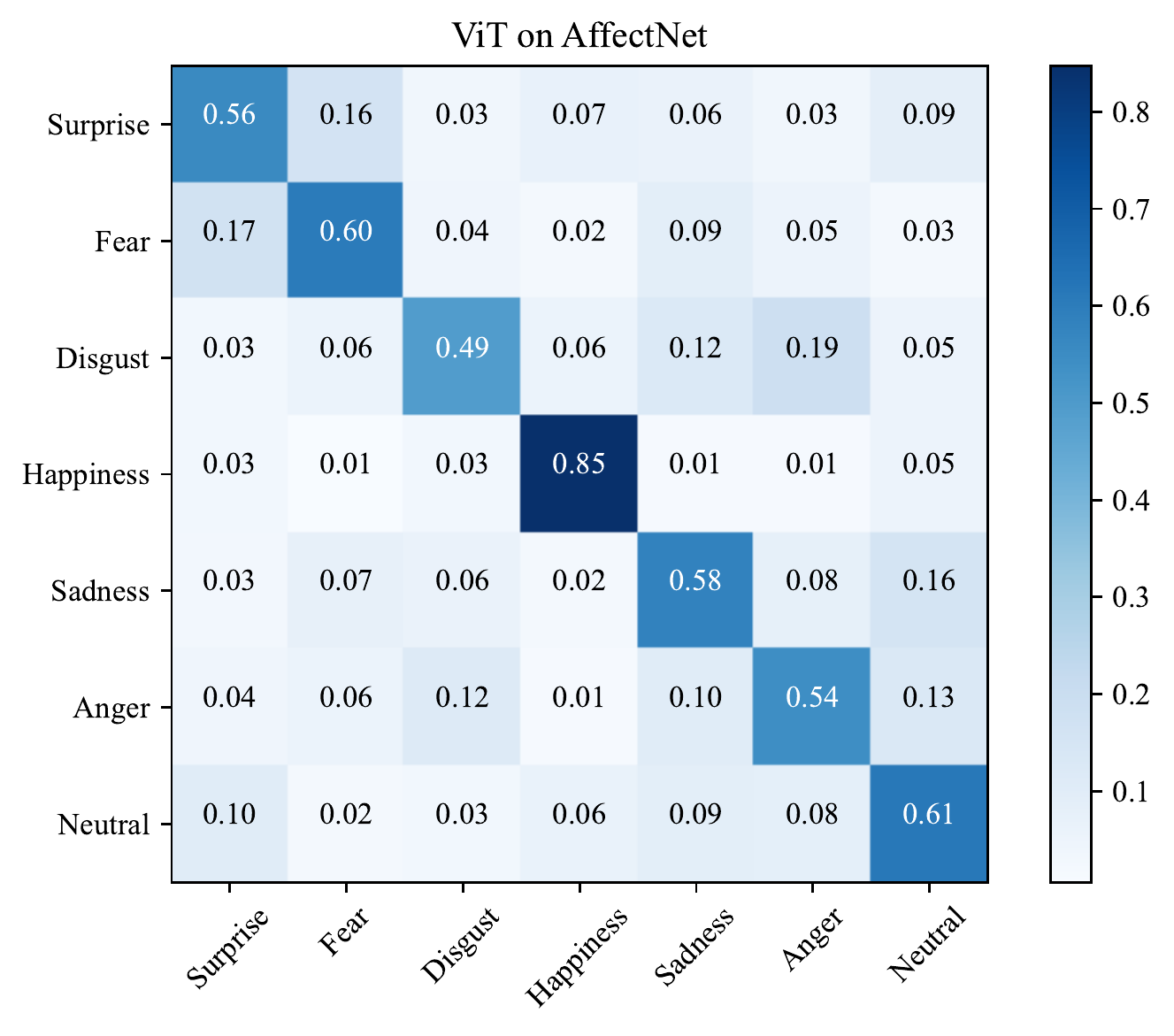}  
\end{subfigure}
\\
\begin{subfigure}{.33\textwidth}
  \centering
  \includegraphics[width=1\linewidth]{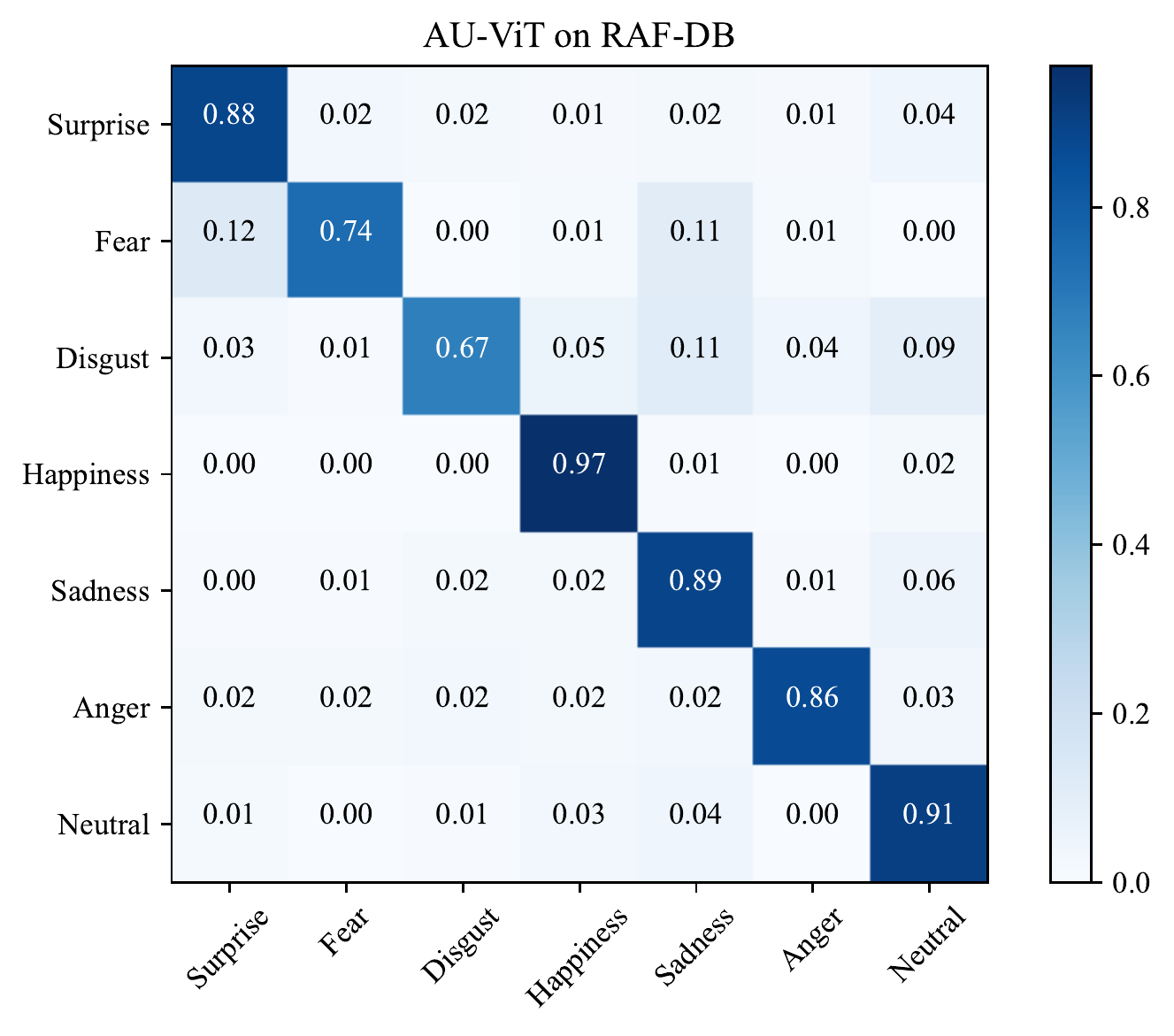}
\end{subfigure}
\begin{subfigure}{.33\textwidth}
  \centering
  \includegraphics[width=1\linewidth]{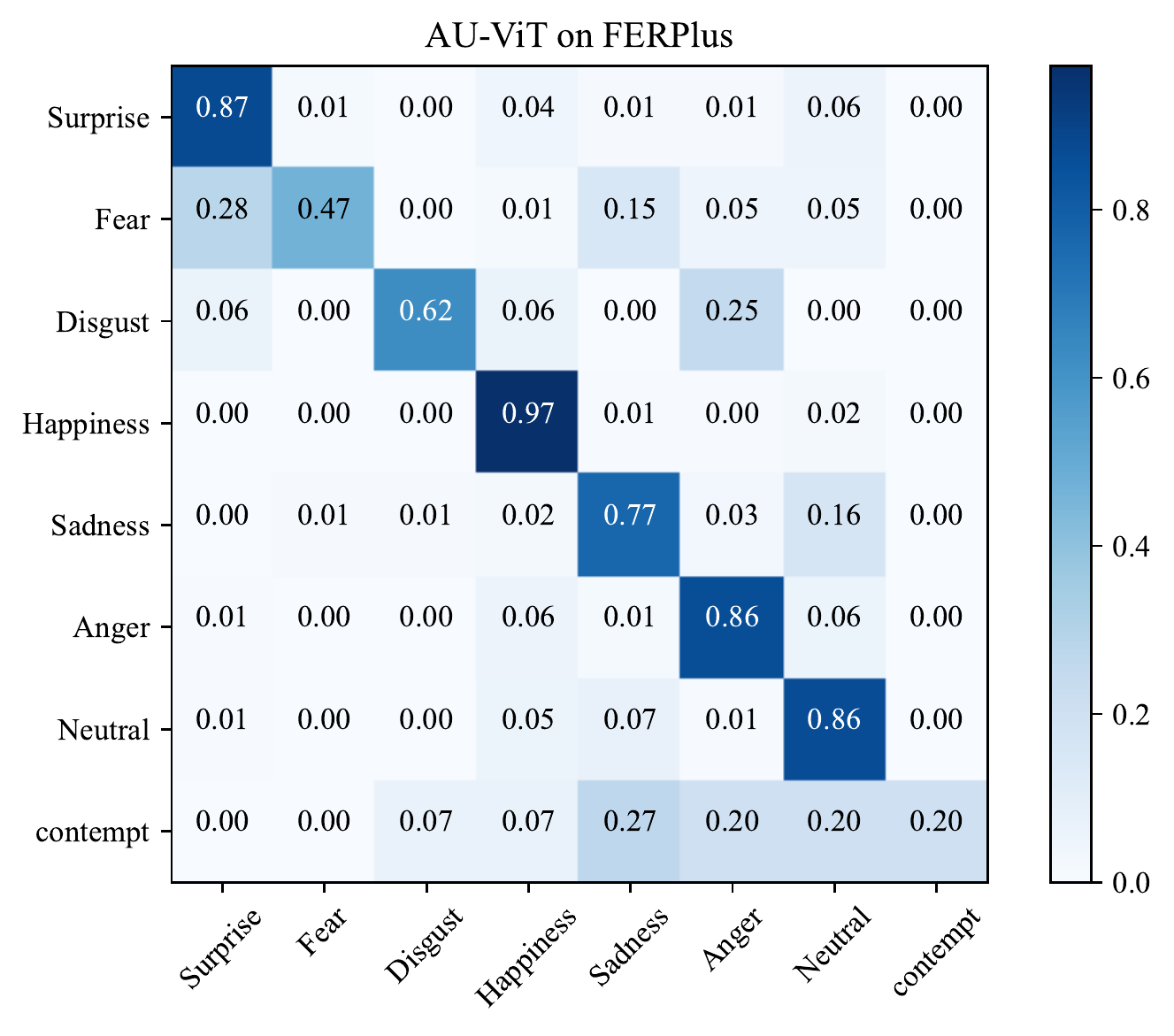}
\end{subfigure}
\begin{subfigure}{.33\textwidth}
  \centering
  \includegraphics[width=1\linewidth]{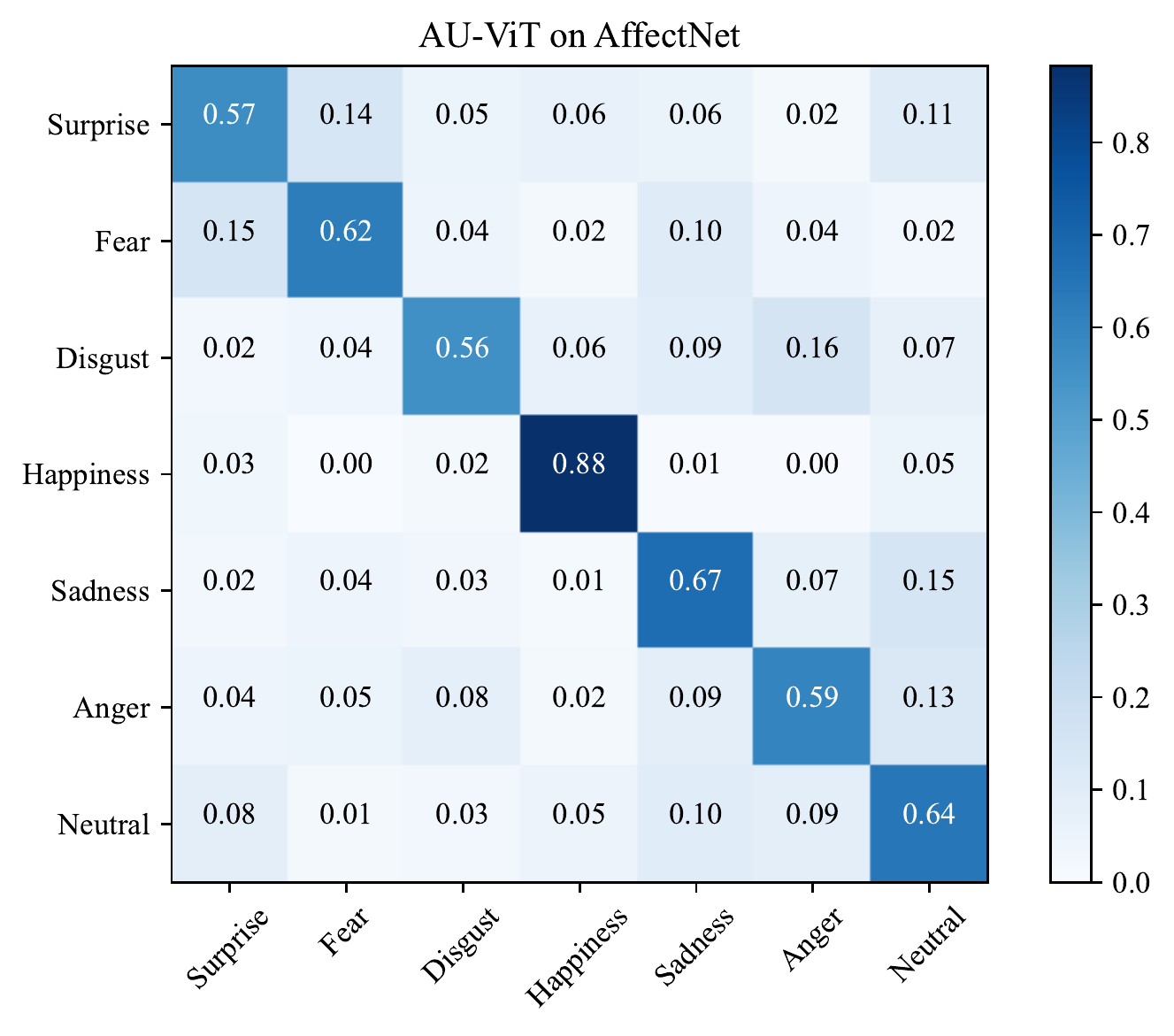}
\end{subfigure}
\caption{The confusion matrices of the vanilla ViT and our AU-ViT on RAF-DB, FERPlus, and AffectNet.}
\label{fig: cm of AU-ViT}
\end{figure*}

\subsection{Datasets}
To evaluate our method, we use three popular in-the-wild facial expression datasets, namely RAF-BD~\cite{li2017reliable}, FERPlus~\cite{BarsoumICMI2016} and AffectNet~\cite{mollahosseini2017affectnet}, and two popular in-the-wild facial AU datasets, namely RAFAU~\cite{Yan_2020_ACCV} and EmotioNet~\cite{fabian2016emotionet}. Besides, we also use the occlusion test datasets~\cite{wang2020region} from RAFDB, FERPlus, and AffectNet.

\textbf{RAFDB}~\cite{li2017reliable}. It consists of 30,000 facial images downloaded from the Internet and annotated with basic or compound expressions by 40 trained human students. It is a large-scale and real-world dataset. In our experiment, we only use images with seven basic expressions (neutral, happiness, surprise, sadness, anger, disgust, fear, neutral), including 12,271 images for training and 3,068 for testing. We mainly report the overall accuracy of the test set.

\textbf{FERPlus}~\cite{BarsoumICMI2016}. The FERPlus contains 28,709 training images, 3,589 validation images, and 3,589 test images collected by the Google search engine. All of the pictures of FERPlus are aligned and resized to 48$\times$48, each of which is annotated by ten annotators. We re-align the faces in the aligned set of FERPlus to increase the area ratio of faces in an image. Apart from seven basic emotions in FER2013, contempt is included in FERPlus. We report the overall accuracy of the test set. 

\textbf{AffectNet}~\cite{mollahosseini2017affectnet}. The AffectNet is the largest dataset that provides both categorical and Valence-Arousal annotations. It includes over one million images from the Internet by querying expression-related keywords in three search engines. In AffectNet, about 420,000 images are manually annotated with eight basic expression labels like RAFDB and 3500 validation images. Note that AffectNet has an imbalanced training set, a test set, and a balanced validation set. So we use an oversampling strategy to produce balanced batch samples in the training phase. The overall accuracy of the validation set is used for measurement.

\textbf{RAFAU}~\cite{Yan_2020_ACCV}. The RAFAU is an extended dataset of RAF-ML collected from the Internet with blended emotions. It varies in subjects' identity, head poses, lighting conditions, and occlusions.
The face images in RAFAU are FACS-coded by two experienced coders independently. It contains 4601 real-world images with 26 kinds of AUs: AU1, AU2, AU4, AU5, AU6, AU7, AU9, AU10, AU12, AU14, AU15, AU16, AU17, AU18, AU20, AU22, AU23, AU24, AU25, AU26, AU27, AU28, AU29, AU35, AU43. In our experiments, we only use the first 21 AUs, which strongly relate to expressions.

\textbf{EmotioNet}~\cite{fabian2016emotionet}. The EmotioNet database includes 950,000 images with annotated AUs. The images are downloaded from the Internet by searching with related facial expression keywords. A specific algorithm automatically annotates these images with AUs, AU intensities, and expression categories. It contains 23 AUs, including AU1, AU2, AU4, AU5, AU6, AU9, AU10, AU12, AU15, AU17, AU18, AU20, AU24, AU25, AU26, AU28, AU43, AU51, AU52, AU53, AU54, AU55, AU56. In our experiments, we only use the first 16 AUs, which strongly relate to facial expressions.

 \textbf{Occlusion FER datasets}~\cite{wang2020region}. The total numbers of occlusion samples in FERPlus (test set), AffectNet (validation set), and RAF-DB (test set) are respectively 605, 682, and 735, which are 16.86\%, 17.05\%, and 23.9\% of their original sets. These real-world test sets are annotated manually with different occlusion types, which include wearing a mask, wearing glasses, objects on the left/right, objects in the upper face and objects in the bottom face, and non-occlusion. Images in the occlusion test sets have at least one type of occlusion. We report the overall accuracy of the three occlusion datasets.
 
\subsection{Joint training with AU-ViT}

We conduct extensive joint training experiments among three FER datasets: RAFDB, FERPlus, and AffectNet, and two AU datasets: RAFAU and EmotioNet. In particular, we predict pseudo-AUs for FER datasets via OpenFace\cite{baltrusaitis2018openface}. The expected AUs include AU1, AU2, AU4, AU5, AU6, AU7, AU9, AU10, AU12, AU14, AU15, AU17, AU20, AU23, AU25, and AU26, which are all used for training. Besides, we select available expression-related AUs if auxiliary AU datasets are used. The specific AUs are mentioned in the datasets section. In the experiments, similar to Table \ref{tab:FER-FER}, the base branch adopts the vanilla ViT and is pre-trained on the VGGFace2~\cite{cao2018vggface2}. The results are shown in Table \ref{tab:FER-AU}.

Compared to joint training with FER labels (i.e., Table \ref{tab:FER-FER}), our AU-ViT significantly enhances the performance, e.g., we achieve a 6.08\% gain for RAFDB when combining RAFDB and RAFAU. These results demonstrate the effectiveness of our AU-ViT. We also find two other observations as follows. First, when combining AU datasets, the gain from RAFAU is generally more significant than that from EmotioNet, which is consistent with the AU distance in Figure \ref{fig: cosine distance}. Second, the auxiliary AU datasets usually bring more improvements, which may be explained that the manual AU annotations have better quality than the pseudo ones. 

To investigate the superiority of our AU-ViT, we show the samples correctly recognized by AU-ViT but not by ViT in Figure \ref{fig: Wrong classed}. From these samples, we find the AU-ViT performs better in four aspects of examples: inter-class similarity, weak expression, occlusion, and profile. Different expressions sometimes share similar facial motions, e.g., a sad face may raise the mouth corner and show the teeth as in a happy face. The ViT trained with FER labels is confused by these samples, while the AU-ViT is not. It may be explained that the AU branch of AU-ViT makes the model pay attention to all the local regions related to action units. Some expressions (e.g., sadness, surprise, and anger) have weak intensity and look similar to neutral. Occlusion and profile faces make it hard for ViT since critical clues, like front faces, are missing. However, our AU-ViT occasionally captures dominated features for expression recognition in non-occlusion facial regions thanks to the AU recognition branch.
We further compare ViT and our basic AU-ViT by illustrating confusion matrices in Figure \ref{fig: cm of AU-ViT}. Our AU-ViT achieves remarkable improvements for the categories `Fear' and `Anger' on RAF-DB, `Disgust' and `Anger' on FERPlus, `Disgust' and `Sadness' on AffectNet.

\begin{table}[t]
\center
\caption{Evaluation of advanced modules in AU-ViT. The results are reported on RAFDB, and the AU-ViT models are jointly trained on RAFDB and RAFAU.}
\label{tab:Ablation_All}
\resizebox{1.0\linewidth}{!}{
\begin{tabular}{@{}cccccc@{}}
\toprule
CNN Embedding   & Multi-stage   & Conv-FFN  & Patch-flatten    & AU Branch & Accuracy  \\
\midrule
                &               &           &                   &           & 87.30     \\
                &               &           &                   &      $\surd$      & 88.80     \\
$\surd$         &               &           &                   &           & 89.86  \\
$\surd$         &$\surd$        &           &                   &           & 90.12 \\
$\surd$         &$\surd$        &$\surd$    &                   &           & 90.22   \\
$\surd$         &$\surd$        &$\surd$    &$\surd$            &           & 90.42  \\
$\surd$         &$\surd$        &$\surd$    &$\surd$            &$\surd$    &\textbf{91.10}  \\
\bottomrule
\end{tabular}}
\end{table}

\begin{figure}[t]
\centering
\includegraphics[width=0.95\linewidth]{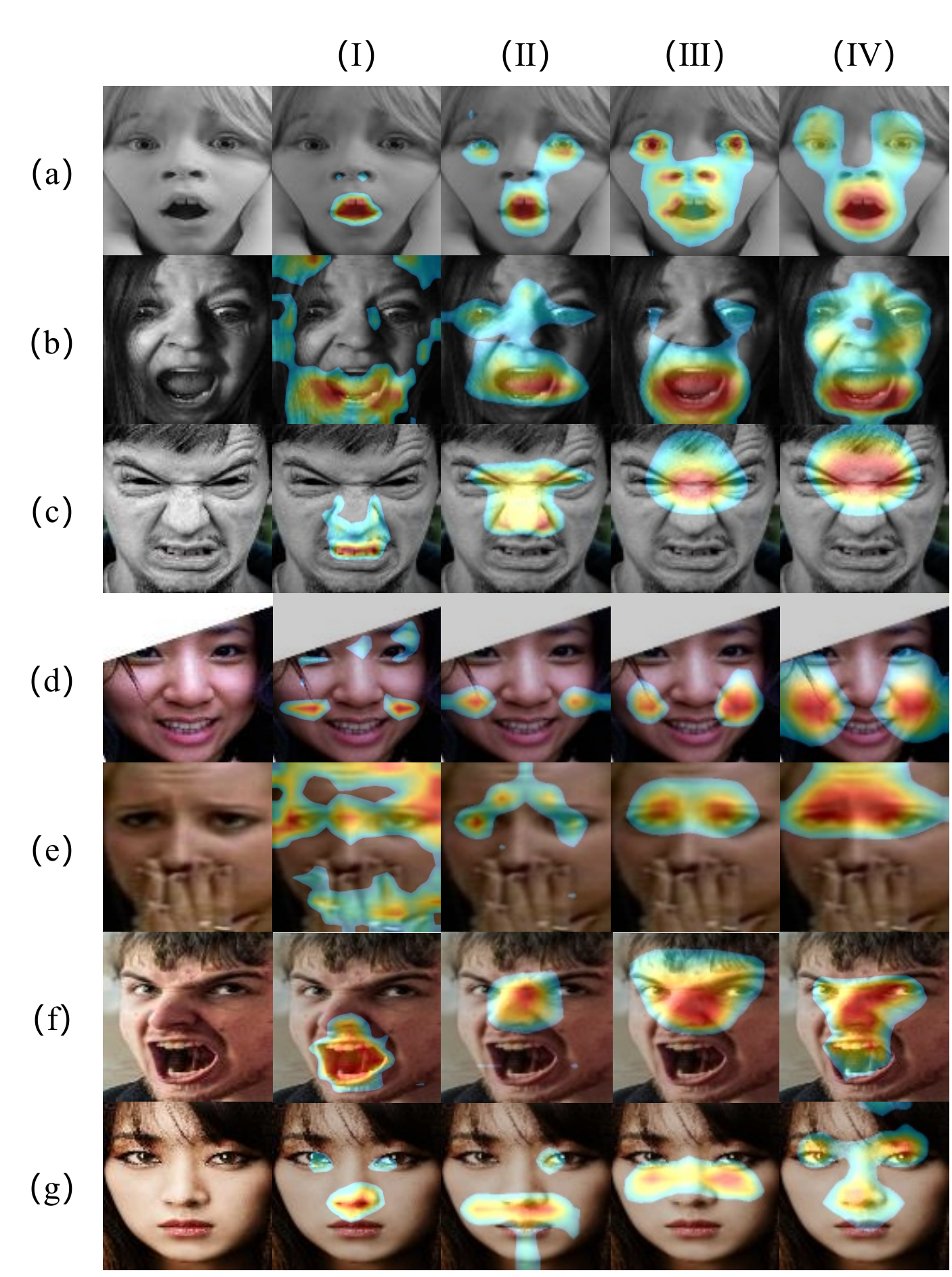}
\caption{Different attention feature maps from (\uppercase\expandafter{\romannumeral1}) the vanilla ViT, (\uppercase\expandafter{\romannumeral2}) ViT with CNN Embedding, (\uppercase\expandafter{\romannumeral3}) ViT with CNN Embedding and Multi-stage blocks, and (\uppercase\expandafter{\romannumeral4}) our final version of AU-ViT. The rows (a)-(g) show seven facial images with different facial expressions from the RAFDB.}
\label{fig:visualization}
\end{figure}

\subsection{Advanced modules for AU-ViT}
Though our basic AU-ViT has already obtained comparable performance to state-of-the-art methods, recently advanced modules for Transformers can further be embedded into our AU-ViT. Thus, an evaluation is designed to investigate the effects of CNN Embedding block, Multi-stage block, Conv-FFN, and Patch-flatten operations. The results are presented in Table \ref{tab:Ablation_All}. For the CNN Embedding block in our base branch, we resort to the IR50 pre-trained on the MS1M dataset~\cite{wang2021face} and use the output feature maps from the third stage as input of Transformer blocks. Adding the CNN Embedding block brings the most significant gain, i.e., 2.56\%, which can be due to the large-scale pre-trained face dataset or the CNN architecture itself. The Multi-stage Transformer block can capture features from more significant regions than standard ViT blocks, which improves the result with CNN Embedding by 0.56\%. We flatten all patch embeddings to keep more information from the last fully-connected layer instead of simply projecting the class token. This simple change also brings 0.2\% performance gain. Note that our critical AU branch consistently obtains visible improvements whether advanced modules are used or not. Our final version of AU-ViT achieves 91.10\% on RAFDB.

\textbf{Visualization}.
To investigate the effects of each module, we employ ScoreCAM~\cite{2020Score} to visualize the attention feature maps from different models in Figure \ref{fig:visualization}. 
From (a) to (g), the categories are surprise, fear, disgust, happiness, sadness, anger, and neutral. The first column shows the aligned input facial images. The second to fifth columns show the results of four different models, namely (\uppercase\expandafter{\romannumeral1}) the vanilla ViT, (\uppercase\expandafter{\romannumeral2}) ViT with CNN Embedding, (\uppercase\expandafter{\romannumeral3}) ViT with CNN Embedding and Multi-stage blocks, and (\uppercase\expandafter{\romannumeral4}) our final version of AU-ViT. 

Comparing different columns, we find that adding CNN Embedding can eliminate unrelated attention regions on the face(e.g., (\uppercase\expandafter{\romannumeral1}), (\uppercase\expandafter{\romannumeral2}) in (b) and (d) ). Multi-stage Transformer guides to learn proper and correct features (e.g. (\uppercase\expandafter{\romannumeral2}), (\uppercase\expandafter{\romannumeral3}) in (c)). The AU branch shrinks the attention maps and focuses on more key areas. Take column (\uppercase\expandafter{\romannumeral4}) in (b) as an example. The model with an AU branch pays attention to the mouth, brow, and eyes, while the model without an AU branch only focuses on the mouth and brow. This visualization can further explain why our AU-ViT outperforms vanilla ViT in the situations of weak expression, occlusion, profile, etc. 

\subsection{Comparison with the state-of-the-art methods}
We compare our final version of AU-ViT to the state-of-the-art methods on RAFDB, FERPlus, and AffectNet.

\begin{table}[tp]
\center
\caption{Comparison to the state-of-the-art results on the FERPlus dataset.}
\begin{tabular}{ccccc}
\toprule
Type                        & Method                        & Year      & Public   & Performance \\\midrule
\multirow{5}{*}{CNN}        &PLD\cite{Barsoum2016TrainingDN}&2016       &ICMI       &85.10 \\
                            &SCN\cite{wang2020suppressing}  &2020       &CVPR       &88.01  \\
                            &RAN\cite{wang2020region}       &2020       &CVPR       &88.55  \\
                            &DMUE\cite{she2021dive}         &2021       &CVPR       &89.51 \\ \midrule
\multirow{3}{*}{Transformer}&VTFF\cite{ma2021facial}        &2021       &TAC        &88.81  \\
                            &MVT\cite{li2021mvt}            &2021       &/          &89.22  \\
                            &Ours                           &2022       &/          &\textbf{90.15}\\
                            \bottomrule
\end{tabular}
\label{tab:sota_FERPlus}
\end{table}

\textbf{Comparison on FERPlus}.
Table \ref{tab:sota_FERPlus} shows the comparison between our AU-ViT and state-of-the-art methods on FERPlus. Among them, PLD~\cite{Barsoum2016TrainingDN}, SCN~\cite{wang2020suppressing} and RAN~\cite{wang2020region} are CNN-based methods. PLD uses VGG13 as the backbone, which is very simple and light. SCN tends to solve the uncertain problem in the FER task. RAN proposes a region-biased loss to encourage high attention weights on the most critical regions.
VTFF~\cite{ma2021facial} and MVT~\cite{li2021mvt} are Transformer-based methods. VTFF also utilizes CNN embeddings and a multi-layer transformer encoder to recognize expressions. MVT uses two Transformer modules to filter out useless patches and classify the remaining features. Our elaborate AU-ViT outperforms these methods and achieves 90.15\% accuracy.

\begin{table}[tp]
\center
\caption{Comparison to the state-of-the-art results on the AffectNet-7 dataset.}
\begin{tabular}{ccccc}
\toprule
Type                        & Method                        & Year      & Public   & Performance \\
\midrule
\multirow{5}{*}{CNN}        &IPA2LT~\cite{zeng2018facial}    &2018       &ECCV       &57.31 \\
                            &LDL-ALSG~\cite{chen2020label}   &2020       &CVPR       &59.35 \\
                            &DDA-Loss~\cite{Farzaneh_2020_CVPR_Workshops} &2020&CVPR &62.34 \\
                            &DMUE~\cite{she2021dive}        &2021    &CVPR           &63.11 \\
                            &EfficientFace~\cite{zhao2021robust} &2021   &AAAI       &63.7 \\
                            &KTN~\cite{2021Adaptively}       &2021       &TIP        &63.97 \\
                            &Meta-Face2Exp~\cite{zeng2022face2exp}  &2022&CVPR       &64.23 \\
                            &DACL~\cite{farzaneh2021facial}  &2021       &WACV       &65.20 \\ 
                            \midrule
\multirow{3}{*}{Transformer}&MVT~\cite{li2021mvt}            &2021       &/          &64.57 \\
                            &VTFF~\cite{ma2021facial}        &2021       &TAC        &64.80 \\
                            &Ours                           &2022       &/          &\textbf{65.59} \\
                            \bottomrule
\end{tabular}
\label{tab:sota_AffectNet}
\end{table}

\textbf{Comparison on AffectNet}.
Table \ref{tab:sota_AffectNet} compares our method to several state-of-the-arts methods on AffectNet.
DDA-Loss~\cite{Farzaneh_2020_CVPR_Workshops} optimizes the embedding space for extreme class imbalance scenarios. 
EfficientFace~\cite{zhao2021robust} focuses on label distribution learning.
KTN~\cite{2021Adaptively} designs a distillation strategy to classify highly similar representations.
DACL~\cite{farzaneh2021facial} focuses on the most discriminative facial regions and features.

Due to the gap between the imbalanced training set and the balanced validation set, we also use the traditional oversampling strategy and increase the proportion of auxiliary data in each iteration from 4:1 to 3:1. We finally obtain 65.59\% on AffectNet, which exceeds the VTFF by 0.79\%. 
\begin{table}[t] 
\center
\caption{Performance comparison with the state-of-the-art methods on RAF-DB.}
\begin{tabular}{ccccc}
\toprule
Type                        & Method                        & Year      & Public   & Performance \\
\midrule
\multirow{5}{*}{CNN}        &DLP-CNN~\cite{2017Reliable}     &2016       &CVPR       & 84.13  \\
                            &LDL-ALSG~\cite{chen2020label}   &2020       &CVPR       & 85.53 \\
                            &IPA2LT~\cite{zeng2018facial}    &2018       &ECCV       & 86.77 \\
                            &RAN~\cite{wang2020region}       &2020       &CVPR       & 86.90 \\
                            &SPDNet~\cite{2018Covariance}     &2018        &CVPRW    & 87.00\\
                            &DDL~\cite{2020Deep}         &2020           &CVPR       & 87.71  \\
                            &SCN~\cite{wang2020suppressing} &2020        &CVPR       & 88.14  \\
                            &Meta-Face2Exp~\cite{zeng2022face2exp}  &2022&CVPR       & 88.54 \\
                            &DMUE~\cite{she2021dive}        &2021    &CVPR           & 89.42 \\
                            &FDRL~\cite{ruan2021feature}    &2021    &CVPR           & 89.72 \\
                            \midrule
\multirow{3}{*}{Transformer}&VTFF~\cite{ma2021facial}        &2021       &TAC        & 88.81 \\
                            &MVT~\cite{li2021mvt}            &2021       &/          & 89.22 \\
                            &TransFER~\cite{xue2021transfer}   &2021    &ICCV       & 90.91\\
                            &Ours                           &2022       &/          &\textbf{91.10}\\
                            \bottomrule
\end{tabular}
\label{tab:sota_RAFDB}
\end{table}

\begin{table}[tp]
\center
\setlength{\tabcolsep}{5mm}
\caption{Comparison between our AU-ViT and other methods with occlusion conditions.}
\begin{tabular}{ccc}

\toprule
Test datasets           & Model     & Acc.    \\\midrule
\multirow{4}{*}{Occlusion-RAFDB}     & ViT(Baseline)  & 81.63 \\
                            & RAN~\cite{wang2020region}        & 82.72  \\
                            & ASF-CVT~\cite{ma2021robust}   & 83.95  \\
                            & AU-ViT    & \textbf{88.02} \\\midrule
\multirow{4}{*}{Occlusion-FERPlus}    & ViT(Baseline)  & 81.48 \\ 
                            & RAN~\cite{wang2020region}        & 83.63  \\
                            & ASF-CVT~\cite{ma2021robust}   & \textbf{84.79}  \\
                            & AU-ViT    & \textbf{84.79} \\\midrule
\multirow{4}{*}{Occlusion-AffectNet}  & ViT(Baseline)  & 57.26 \\ 
                            & RAN~\cite{wang2020region}       & 58.50  \\
                            & ASF-CVT~\cite{ma2021robust}   & 62.98  \\
                            & AU-ViT    & \textbf{63.92} \\\bottomrule
\end{tabular}
\label{tab:occlusion}
\end{table}

\textbf{Comparison on RAFDB}.
Table \ref{tab:sota_RAFDB} compares our approach with previous state-of-the-art methods on RAFDB. 
Among all the competing methods, DDL~\cite{2020Deep} and RAN pay attention to disentangling the disturbing factors in facial expression images. SPDNet~\cite{2018Covariance} introduces a new network architecture, and DLP-CNN~\cite{2017Reliable} uses a novel loss function to explore intra-class variations. IPA2LT~\cite{zeng2018facial} and SCN~\cite{wang2020suppressing} try to reduce the effects of noise labels. TransFER~\cite{xue2021transfer} is the first Transformer-based model for the FER task. It is worth noting that IPA2LT also resorts to multiple facial datasets for performance boosting via multiple predicted pseudo-FER labels from varied models. 
All the above methods improve FER performance by focusing on facial expression images, model architectures, and FER labels, neglecting the importance of AUs. Our AU-ViT finally achieves 91.10\% on RAFDB, which surpasses TransFER slightly, establishing a new state of the art.

\textbf{Comparison on occlusion datasets}
As shown in Figure \ref{fig: Wrong classed}, our AU-ViT works better than ViT in occlusion and profile faces, thanks to the guidance of the AU branch. To further verify this, we evaluate our basic AU-ViT (without advanced modules) on the occlusion test sets of RAFDB, FERPlus, and AffectNet. All the models are pre-trained on VGGFace2~\cite{cao2018vggface2} and fine-tuned on the corresponding training set. The evaluation results are shown in Table~\ref{tab:occlusion}.

Two observations can be concluded as follows. First, our AU-ViT improves the baseline ViT on all datasets, with gains of 6.39\%, 3.31\%, and 6.66\% on Occlusion-RAFDB, Occlusion-FERPlus, and Occlusion-AffectNet, respectively. Second, though ASF-CVT utilizes Transformers on CNN feature maps, our basic AU-ViT outperforms it on Occlusion-RAFDB and Occlusion-AffectNet with large margins. 
Worth mentioning that the symmetric maxout layer in the AU branch can effectively capture AU activation even if half of the face is occluded. We experimentally find that the symmetric maxout layer can consistently bring around 0.5\% performance gain.

\begin{table*}[t]
\center
\caption{Performance comparison between the AU Token and the AU branch on RAFDB.}
\begin{tabular}{c|ccccccc}
\hline
\diagbox{Target}{Accuracy (\%)}{Auxiliary}                     &  &RAFDB  &FERPlus& AffectNet   &RAFAU & EmotioNet &Baseline \\\hline
\midrule
\multirow{2}{*}{RAFDB} &AU token  &87.35  &87.71      & 87.65     &87.97  &87.03 &\multirow{2}{*}{87.3}\\
                                           &AU branch &87.81  &88.41      & 87.84     &88.8   &87.9  &\\
\bottomrule
\end{tabular}
\label{tab:Token_vs_Branch}
\end{table*}

\begin{figure}
    \centering
    \includegraphics[width=1\linewidth]{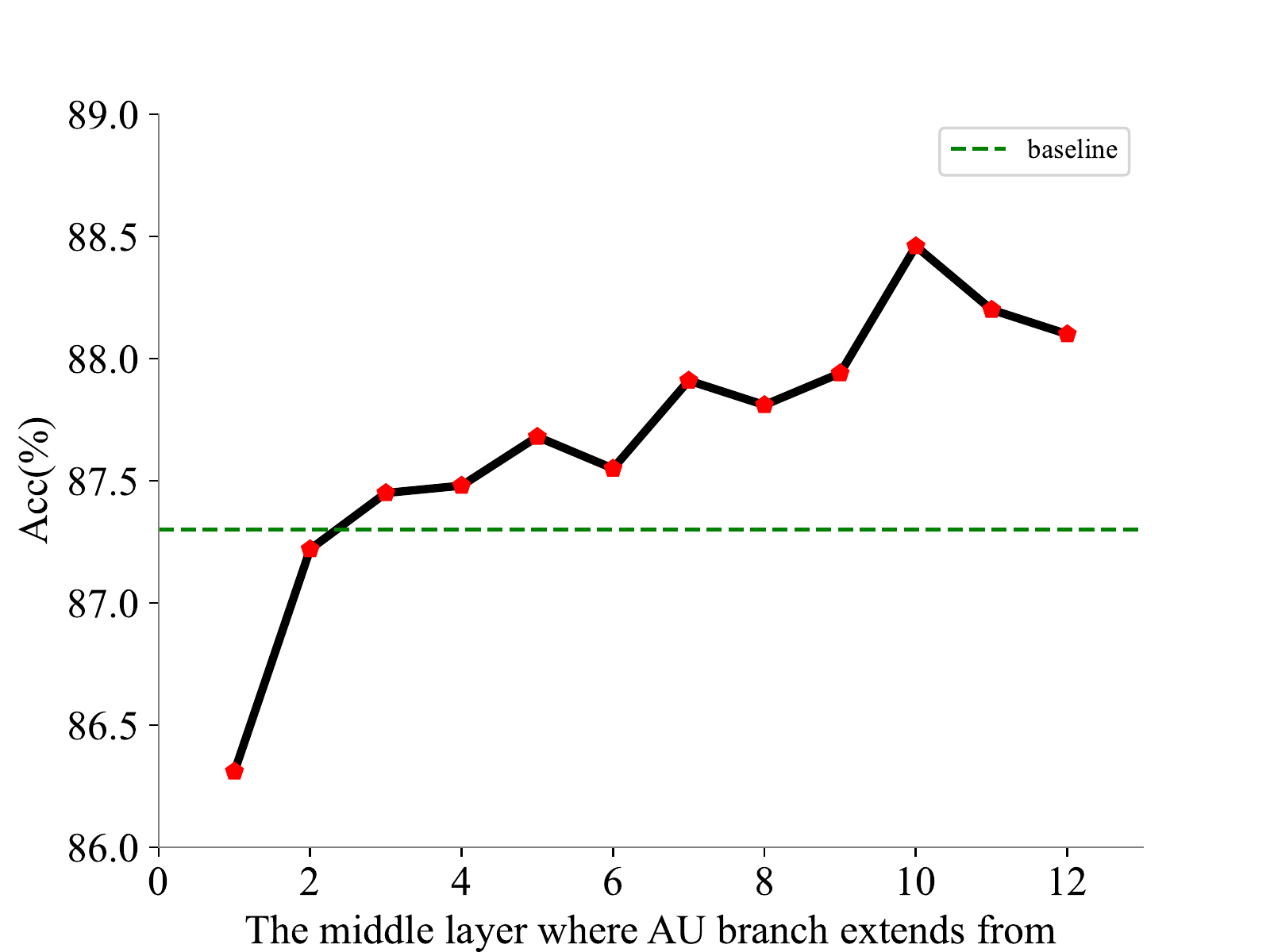} 
    \caption{Evaluation of the AU branch position in our basic AU-ViT.}
    \label{fig:position}
\end{figure}

\begin{figure}
    \centering
    \includegraphics[width=1\linewidth]{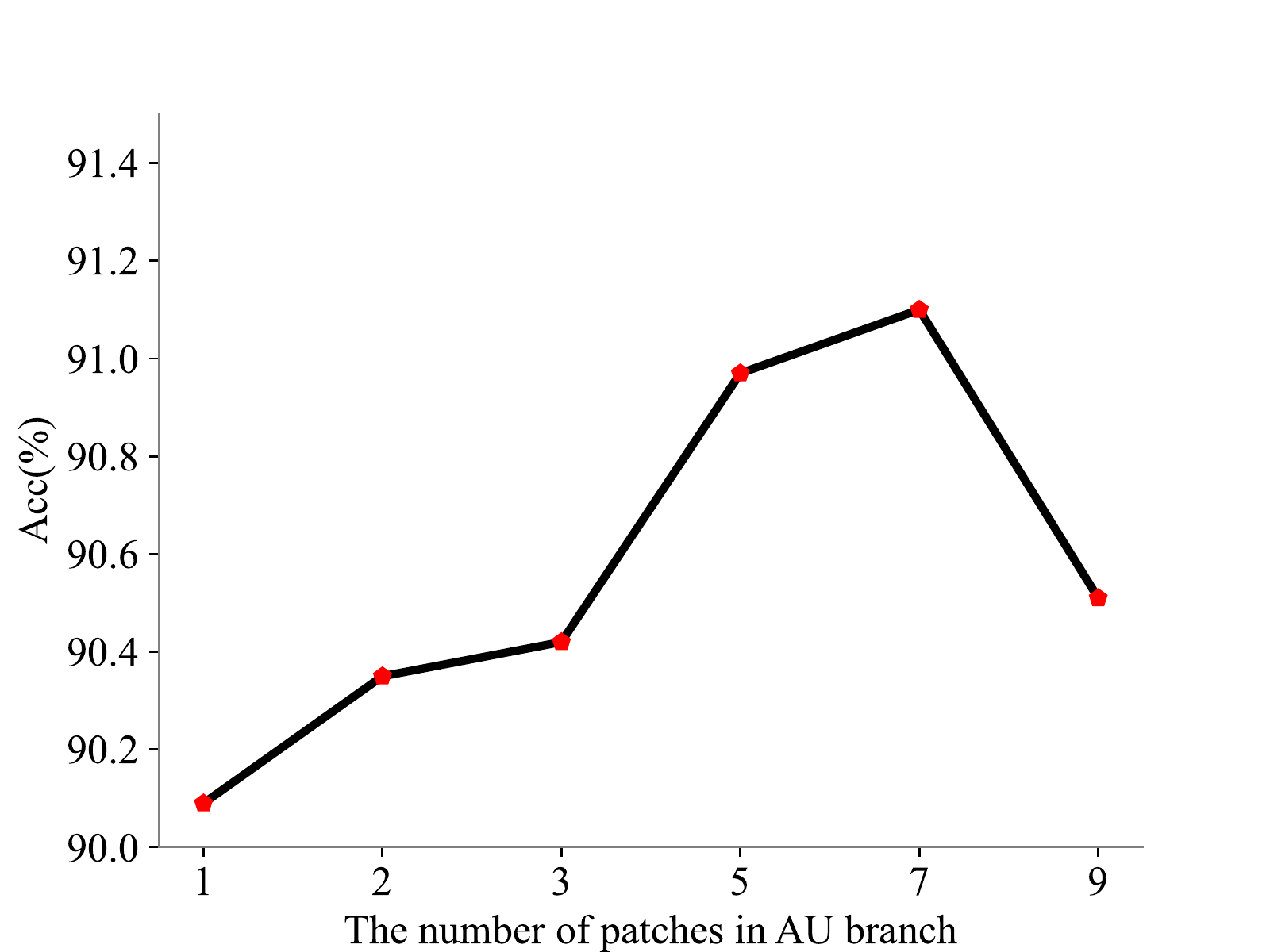} 
    \caption{Evaluation of the splitting strategy in AU branch of our final AU-ViT.}
    \label{fig:split}
\end{figure}

\begin{figure*}[t]
\centering
\includegraphics[width=0.9\textwidth]{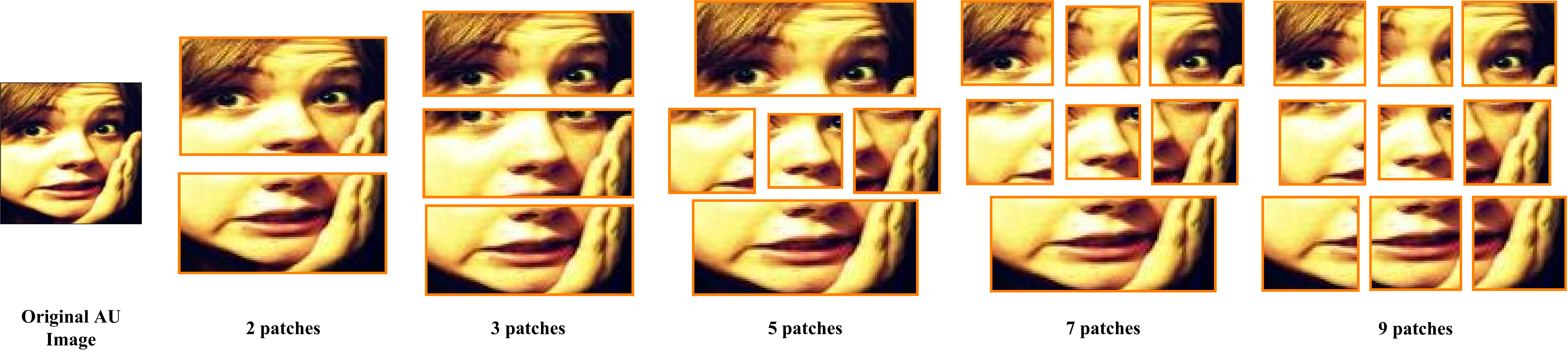}
\caption{Evaluation of different patch-splitting strategies.}
\label{fig:patch splitting}
\end{figure*}

\subsection{Ablation study}
Adding AU information in FER training is previously used for cross-domain facial expression recognition~\cite{wang2020guided} or compound expression recognition~\cite{pu2021expression}. Indeed, it is not trivial to boost performance on individual FER datasets with AUs. Here, we make ablation experiments to show our efforts on this issue.

\textbf{The usage of AUs}.
With AUs, the simple idea is to improve a FER model with a multi-task training strategy. Following the idea, we add an AU Token in the baseline ViT and conduct multi-task training on RAFDB with other auxiliary datasets. Specifically, we concatenate the AU token with patch embeddings and the class Token after the linear projection. We implement a multi-layer perception on the AU Token to estimate AUs, and the others are the same as baseline ViT. Table \ref{tab:Token_vs_Branch} compares these two usage of AUs. It is worth noting that we can also use the pseudo-AUs of RAFDB for training. From Table \ref{tab:Token_vs_Branch}, several observations can be concluded as follows. First, compared to the baseline, the `AU token' multi-task training method can also slightly improve performance except for training with EmotioNet. Second, our AU branch consistently improves the baseline with visible margins. Third, our AU branch not only boosts performance with all auxiliary datasets but also outperforms the 'AU token' strategy. Thus, we believe that our elaborately-designed AU branch with patch splitting and symmetric maxout can be more precise than just adding an AU token for AU estimation.

\textbf{The position of AU branch.}
In our AU-ViT model, the base and Exp-ViT branches own 12 Transformer blocks in succession, and our AU branch is extended from the default 10-th block. To investigate the effect of AU branch position, we evaluate RAFDB with auxiliary RAFAU by setting the position from 1 to 12 and show the results in Figure \ref{fig:position}. Adding the AU branch in early blocks can be harmful, which may be explained by the fact that the early blocks do not have enough semantic information for AU estimation, thus disturbing the low-level feature learning for FER. As the AU branch position goes deep, the performance is gradually improved and saturated at the default position. Extending from the last two blocks may slightly damage the semantic features of FER, leading to performance degradation.

\textbf{The patch splitting strategy.}
Since AUs are defined as the movement of muscles in the face, they are naturally related to local regions. We believe that splitting a face image into different patches helps recognize AUs. However, which splitting scheme is the best for the final performance is unknown. To this end, we design five strategies for our default AU branch, as shown in Figure \ref{fig:patch splitting}. The splitting schemes generate 2, 3, 5, 7, and 9 regions for each face image. We do not split the bottom half face for most schemes since 14 AUs are related to the lower part of the face, and it isn't easy to separate individual AUs. All the patches are slightly overlapped for the necessity of complete AU-related features. 

We evaluate with our final version of AU-ViT and present the results in Figure \ref{fig:split}. Without splitting (i.e., one patch), our AU-ViT obtains 90.09\%. It slightly improves performance by dividing it into 2 or 3 patches in a row. Further separating the upper face in row and column dramatically boost the accuracy. We get the best performance by the default 7-patch splitting strategy. We observe a visible performance degradation when further splitting the bottom face. All in all, breaking the feature maps into multiple patches keeps spatial information and benefits AU estimation and the final FER performance.


\section{Conclusion}
In this paper, we first introduce the expression-special mean images and mean activated AU images to show the FER dataset bias, which are consistent with the quantitative results of joint training experiments. Meanwhile, we present an AU-aware ViT model for facial expression recognition in the wild. It mainly resorts to the AU information, and effectively boosts the performance of individual FER datasets by leveraging other biased datasets. We carefully design the AU-ViT model with several advanced modules and observe that our model can capture more local region features than traditional ViT. We finally achieve state-of-the-art performance in most of the popular FER datasets including occlusion datasets. 



%
\IEEEpeerreviewmaketitle

\bibliographystyle{plain}

\bibliography{egbib}



%







\end{document}